\documentclass[11pt]{article}

\usepackage[final]{acl}

\usepackage{times}
\usepackage{latexsym}

\usepackage{microtype}
\usepackage{hyperref}

\usepackage{inconsolata}

\usepackage[inline]{enumitem}

\usepackage{graphicx}
\usepackage{xspace}

\usepackage{amsmath}
\usepackage{multicol}
\usepackage{graphicx}
\usepackage{booktabs}
\usepackage{array} 
\usepackage{amssymb}
\usepackage{tabularx}
\usepackage[table]{xcolor}
\usepackage{colortbl}
\usepackage{float}
\usepackage{placeins}
\usepackage{needspace}
\usepackage{capt-of}

\definecolor{headerblue}{RGB}{41, 67, 92}
\definecolor{rowaccent}{RGB}{245, 247, 250}

\usepackage[T2A,T1]{fontenc}   %
\usepackage[utf8]{inputenc}    %
   \usepackage[german,russian,english]{babel}
\shorthandoff{"}

\title{Voices of Freelance Professional Writers on AI: \\ Limitations, Expectations, and Fears}

\author{
    \begin{minipage}[t]{\textwidth}
        \centering
        \normalfont
  \textbf{Anastasiia Ivanova}$^{1}$\thanks{ Correspondence: \texttt{anastasiia.ivanova@campus.lmu.de}} \,
  \textbf{Natalia Fedorova}$^{2}$ \
 \textbf{Ekaterina Artemova}$^{3}$ \\\ 
        {
            \textsuperscript{1} LMU Munich
            \textsuperscript{2} Independent researcher 
            \textsuperscript{3} Toloka AI
        }
    \end{minipage}
}

\newcommand{\toloka}{\texttt{Toloka}\xspace}

\begin{document}
\maketitle

\begin{abstract}
The rapid development of AI-driven tools, particularly large language models (LLMs), is reshaping professional writing. Still, key aspects of their adoption such as language support, ethics, and long-term impact on writers' voice and creativity remain underexplored. In this work, we carried out a questionnaire ($N = 301$) and an interactive task ($N = 36$) targeting freelance professional writers regularly using AI. We examined AI-assisted writing practices across 25+ languages, ethical concerns, and user expectations. Our findings reveal that adoption is shaped more by peer influence and professional outlook than demographics, that multilingual writers face both performance and perception barriers to equitable AI use, and that job security concerns are widespread and adoption-independent.
\end{abstract}

\section{Introduction} \label{sec:introduction}
\begin{figure*}
    \centering
    \includegraphics[width=\textwidth]{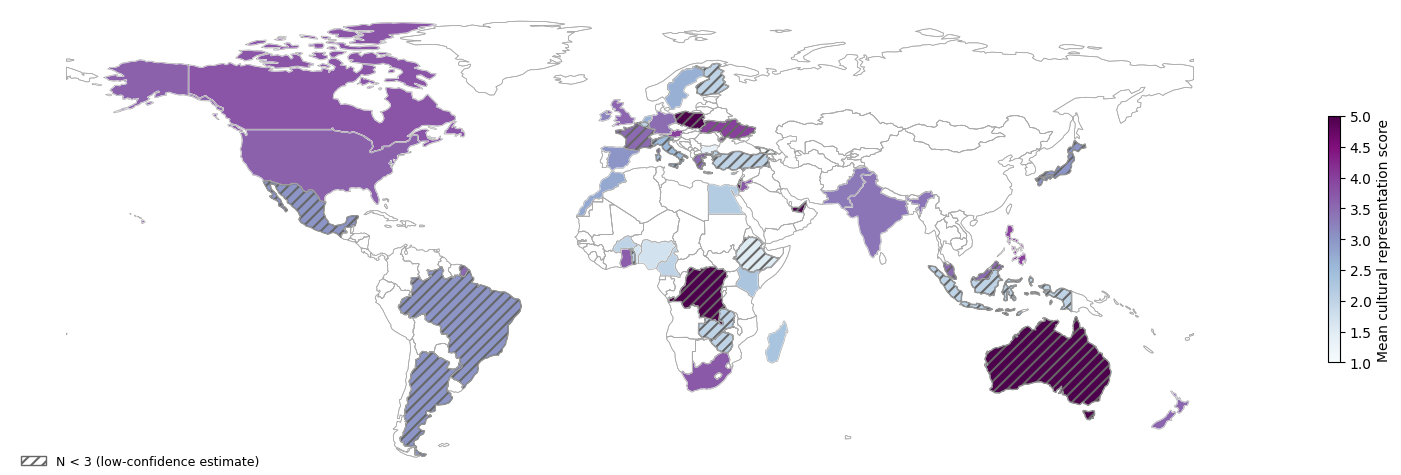}
\caption{Perceived cultural representation in AI chatbots (N = 301). Responses to ``Do you feel your culture is adequately represented in AI chatbots, regardless of the language used?'' Higher scores indicate greater representation. Countries with fewer than 3 respondents are marked with diagonal hatching and should be read as individual data points rather than stable per-country estimates; white indicates no respondents from that country.}
    \label{fig:map}
\end{figure*}
\setlength{\textfloatsep}{8pt}

Recent advancements in large language models (LLMs) have led to their rapid adoption across variety of professional occupations. While integration of LLM-driven AI chatbots has been studied in literature for areas such as research writing \citep{liao2024llms}, software engineering \citep{stackoverflow2024}, and creative industries \citep{guo2024pen}, less attention has been paid to their adoption in freelance professional writing. The latter can include copywriting, content creation for social media, or editing and assumes strong writing skills and expertise in a specific genre or domain \citep{kellogg2018}. Recently AI chatbots found their place among specialized tools to enhance the productivity of writers and the quality of their grammar and style. Still, their usefulness in practice is underexplored.

To address this gap, we present what is, to our knowledge, the largest multilingual study of professional writers' experiences with AI chatbots, combining a large-scale questionnaire with a behavioral writing task measuring human intervention in AI-generated text. Throughout this paper, \textit{professional writers} refers to writers whose primary income derives from producing texts for clients or platforms. Our participants are writers on a freelance writers platform \toloka{} whose primary work involves producing comprehensive texts, e.g., instruction-tuning datasets, question-answering corpora, and post-training data.

Our study combines two complementary instruments: a questionnaire (N = 301 across 25+ languages from 49 countries;~\autoref{fig:map}) capturing writers’ practices, language-specific challenges, expectations, limitations, and ethical concerns; and an interactive writing task (N = 36) measuring editing decisions and prompting strategies directly. We address three research questions:

\begin{itemize}[leftmargin=2.5em, labelsep=0.5em, topsep=0pt, partopsep=0pt, itemsep=0pt, parsep=0pt]
\item[\textbf{RQ1}] How do freelance writers integrate AI chatbots into practice, and how does this shape their writing experience and productivity?
\item[\textbf{RQ2}] How effective are AI chatbots in non-English languages, and do they represent diverse cultural knowledge?
\item[\textbf{RQ3}] What ethical and professional concerns do professional writers hold toward AI?

\end{itemize}

We find that, first, adoption (73\%) is driven more by peer influence and creative outlook than demographics, a pattern consistent across all three RQs. Second, multilingual writers show limited awareness of available tools, rely mostly on ChatGPT, and refrain from using languages they incorrectly believe are unsupported. Writers from non-Western backgrounds consistently report  underrepresentation of their culture. Third, job security concerns are widespread and adoption-independent, influenced by outlook and usage patterns rather than demographics. These findings have direct implications on how multilingual LLMs are evaluated, documented, and deployed.\footnote{The project data are available in \href{https://github.com/Toloka/surveying-prof-writers-on-ai}{the project repository}.}

\section{Related work} \label{sec:related_work}
\noindent \textbf{Human-AI Collaboration in Writing.} Prior work on human-AI collaboration in writing focused primarily on academic and creative contexts, leaving professional freelance writing understudied. Studies of academic writers document common use cases from proofreading to brainstorming and identify diverse prompting strategies \citep{fecher2023friend, morris2023scientists, liao2024llms}. Large-scale log analysis of Bing Copilot and WildChat sessions identifies eight prototypical collaboration behaviors (revising intents, exploring alternatives, injecting content etc.) that account for 80–85\% of variation in user-LLM interaction \citep{mysore2025prototypical}; our interactive task states iterative refinement as the dominant strategy among professional writers, but extends this to multilingual settings where cultural constraints add a layer of revision. Among creative writers, \citet{guo2025pen} report strong tensions around authenticity and creative identity; we partially replicate concerns about voice and creative identity but observe lower resistance to adoption, likely reflecting differences between artistic and commercial writing contexts. \citet{lamb2024precaritysolidaritypreliminaryresults} find that queer and disabled fiction writers are markedly more opposed to AI than non-marginalized peers; that study frames adoption as an ethical problem within those communities. These findings suggest that professional context may shape AI acceptance differently from artistic communities.

\noindent \textbf{Authorship, Ownership, Professional Identity.} Writers consistently struggle to attribute authorship in AI-assisted work. \citet{hwang2024} find that professional writers negotiate partial attribution ("80\% me, 20\% AI") and resist publicly declaring AI co-authorship. This raises open questions about how commercial writing contexts negotiate authorship and disclosure. \citet{li2024value} show experimentally that content-generation assistance significantly reduces perceived ownership and pride in writing outcomes; our results qualify this picture. Heavy professional users maintain ownership claims despite daily AI use, possibly because they retain strong editorial control. \citet{draxler2024ai} document the ``AI Ghostwriter Effect'' -- claiming authorship without perceiving ownership -- which aligns with the authorship ambiguity we observe, though writers in our sample more frequently resolve this ambiguity toward full authorship rather than uncertainty. Beyond ownership, \citet{hu2025negotiating} find that non-native English academic writers describe five different ways people perceive and treat ChatGPT; our multilingual professional writers show similar variation, but commercial rather than scholarly stakes shape the negotiation.

\noindent \textbf{Multilingual AI: Performance, Perception, and Culture.} Multilingual LLMs exhibit fluency in non-English languages but frequently fail at cultural context, idioms, and factual accuracy \citep{huang2024survey, pawar2024survey, myung2024blend}, stemming in part from English-dominant training \citep{shaib-etal-2024-detection, bender2021dangers}. Models perform best when language and cultural context align that breaks down for lower-resource languages \citep{ying2025disentangling}. User-facing evaluations reveal Western-centric calibration: alignment with cultural values decreases with cultural distance from the US \citep{van2026cultural}, and models consistently underrepresent religious and value-based priorities of Asian minority groups \citep{biswas2025mind}. Preference learning with native cultural data improves alignment but leaves persistent gaps for under-resourced regions \citep{guo2025care}. 

Prior work focuses primarily on model-side properties such as multilingual performance and cultural alignment. Less attention has been paid to user-side perceptions of language support and whether mismatches between actual and perceived capability influence adoption -- a potential perception gap in multilingual AI use.

\section{Methodology} \label{sec:methodology}
\subsection{Participants} %

All participants are verified freelance writers on \toloka, a platform for high-quality text production for generative AI applications. Unlike standard crowdsourcing (simple labeling, classification), writing tasks on \toloka require producing complete, grammatically correct texts -- for example, instruction-tuning data for a conversational agent in an online marketplace. Before joining, all writers pass a platform-administered selection process evaluating professional experience and relevant education. The most common daily tasks reported in the questionnaire included writing, editing, proofreading, researching new content, communication with teams and management, content performance analysis, and translation, confirming participants' primary engagement in professional text production.

Questionnaire participants (N=301) were invited via email based on multilingual proficiency, experience with open-ended writing tasks, and relevant professional background; approximately 40\% of invited writers completed it. A subset was then invited to the interactive task based on: (i) self-reported proficiency in non-English languages, (ii) low ratings of AI performance in those languages, and/or (iii) detailed free-form comments about AI behavior. Of ~120 invited, 36 completed the task (30\%). Participation was voluntary and compensated at established hourly rates; opening the survey constituted consent (see Sec. \ref{data_collection}), and participants were informed about study purposes and agreed to the anonymized publication of their responses.

\textbf{Participant background.} Participants were born in 49 countries (\autoref{fig:map}) and showed a near-equal gender distribution (52\% men, 47\% women). Most held either an undergraduate degree (37\%) or a master's degree (40\%) as their highest level of education. Age and professional experience distributions are shown in \autoref{fig:age-exp-stat} and \autoref{appendix:statistics}. Additional anonymized participant information is provided in the supplementary materials.

\subsection{Questionnaire Design} 

Our questionnaire is inspired by prior research exploring attitudes toward AI in professional and language communities \citep{liao2024llms,stackoverflow2024,guo2024pen,blaschke-etal-2024-dialect}. To better position the questions and multiple choice answers options, we conducted pilot video interviews with five volunteer writers spanning diverse linguistic backgrounds (Zulu, Brazilian Portuguese, Hindi, Filipino, and Afrikaans), professional fields (media studies, philology, and others), and age ranges, each with at least ten years of writing or editing experience. Their feedback formed an initial draft, which was then reviewed by 10 additional freelance writers outside the study sample for further refinement.

The final questionnaire consisted of 64 questions (\autoref{appendix:survey}), including five open-ended items and eight with an open ``Other'' response option, 
organized as:

\begin{enumerate}[noitemsep, topsep=0pt]
\item \textbf{Demographics and basic information}: Country of birth and residence, age, gender, education, preferred languages of communication, and work experience.
\item  \textbf{AI chatbot usage patterns}: Frequency of use, preferred AI chatbots, prompting strategies, tasks performed, and willingness to pay for using AI chatbots.
\item  \textbf{AI chatbots in writing tasks:} Main advantages and disadvantages, influence on writing style and voice, and ability to recognize machine-generated texts.
\item \textbf{Non-English language experience}: Language proficiency and use at work, frequency and tasks in non-English languages, performance evaluation, and reasons for non-use where applicable.
\item  \textbf{Cultural knowledge}: Perception of cultural representation in AI chatbot outputs.
\item  \textbf{Limitations and outlook}: Perceived limitations, expectations for future developments, and changes in perception over time.
\item \textbf{AI impact and ethical issues}: Job displacement and plagiarism, sentiments toward AI progress, sense of ownership and authorship in collaborative work with AI chatbots.

\end{enumerate}

\subsection{Data Analysis}
\label{sec:analysis}
We encoded non-numeric variables as ordinal when a natural order existed in the answer options (e.g., age intervals) or as one-hot vectors when no such order was present (e.g., methods of learning to use AI chatbots). Countries of birth and residence were encoded based on their frequency in our sample. After encoding all data and removing zero-variance features, we obtained 75 variables for analysis.

Factor analysis was performed on the 220 respondents who reported using AI chatbots, as factor-related questions were only applicable to users. We used maximum likelihood estimation (\texttt{sklearn.decomposition.FactorAnalysis}) on non-demographic and non-target variables to capture underlying attitude patterns. Missing values were imputed using mean substitution. The optimal number of factors was determined through PCA scree plot analysis, yielding seven factors that collectively explained 21.4\% of the variance. Factors were interpreted based on variables with absolute loadings greater than 0.4, following established guidelines \citep{hair2013multivariate}. These factor scores were then used as predictor variables in linear regression analyses to address our research questions. Factor analysis was preferred over pairwise correlation approaches because preliminary analysis found no interpretable pairwise correlations after multiple comparison correction. For instance, age, gender, and education did not predict adoption or usage frequency indicating that variance is distributed across latent attitudinal dimensions rather than concentrated in individual variables; the factor structure's convergence with qualitative findings from the interactive task provides additional cross-instrument validation.

Relationships between variables were examined using appropriate statistical methods based on variable types. Spearman's rank correlation coefficient \citep{spearman1961proof} was applied for ordinal variables, chi-square ($\chi^2$) tests \citep{pearson1900x} for binary variables, and point-biserial correlation \citep{maccallum2002practice} for binary-ordinal relationships. Statistical analyses were implemented using \texttt{scikit-learn}~\cite{scikit-learn}. All reported $p$-values are corrected for multiple comparisons using the Benjamini-Hochberg procedure. 

Qualitative comments were analyzed thematically following \citet{braun2006} to identify recurring user perceptions, operational workflows, and socio-technical concerns regarding LLM usage. Two annotators independently reviewed the responses to generate an initial codebook through iterative inductive coding. Codes were subsequently synthesized into four high-level themes and ten granular subthemes capturing linguistic, technical, and ethical perspectives. Disagreements in code assignment were resolved through consensus discussions to ensure thematic coherence. The complete qualitative codebook, including subtheme definitions, scope descriptions, and corresponding survey quote mappings, is detailed in Appendix~\ref{app:codebook} (Table~\ref{tab:qualitative_codebook}).

\subsection{Interactive Task} 

The interactive task (materials in \autoref{appendix:interactive_task}) complements RQ2's self-reported multilingual usage patterns with behavioral evidence. Self-reported performance ratings can be shaped by recency, emotional salience, or beliefs about a language's prestige, independent of actual model behavior; the interactive task addresses this by measuring concrete edit behavior on a shared, controlled task rather than relying on retrospective self-assessment. The task was designed to observe writers' prompting strategies and measure human intervention in AI-generated text in non-English languages. Participants created a local event advertisement using their preferred AI chatbot and refined its output. The advertisement genre was chosen deliberately for its cultural specificity -- effective local promotion requires cultural knowledge, idiomatic language, and audience awareness that generic writing tasks do not elicit. Participants submitted their full interaction log, the AI-generated version selected for editing, the final advertisement, and written comments on revisions, particularly highlighting cultural or linguistic issues encountered.

To quantify human improvements, final texts were compared to the original AI outputs using the \texttt{SequenceMatcher} similarity ratio\footnote{An algorithm from the \texttt{difflib} Python library (\href{https://docs.python.org/3/library/difflib.html}{docs.python.org/3/library/difflib.html})} (see \autoref{fig:inter_task_example} for an example), yielding a score from 0 (completely different) to 1 (identical). As this metric relies on surface-level string matching, scores may be inflated for semantically significant edits. We treat them as a conservative lower bound on human intervention.

\subsection{Data Collection and Ethics} \label{data_collection}

The questionnaire and interactive task were hosted on \toloka,  with estimated completion times of $\sim$45 and $\sim$60 respectively, based on mean times of early respondents. All participants were compensated at their established hourly rates.

Mandatory questions covered all sections except the non-English language block (shown only to non-English speakers) and open-ended items. All interactive task questions were mandatory.

Participation was voluntary; opening the survey constituted informed consent. No IRB approval was required, as the study was conducted as a standard paid task within \toloka's workflow, covered by participants' user agreement and privacy notice. Relevant excerpts are in \autoref{appendix:ethics}.

\section{Results} \label{sec:results}

\begin{figure*}
    \centering
    \includegraphics[width=\textwidth]{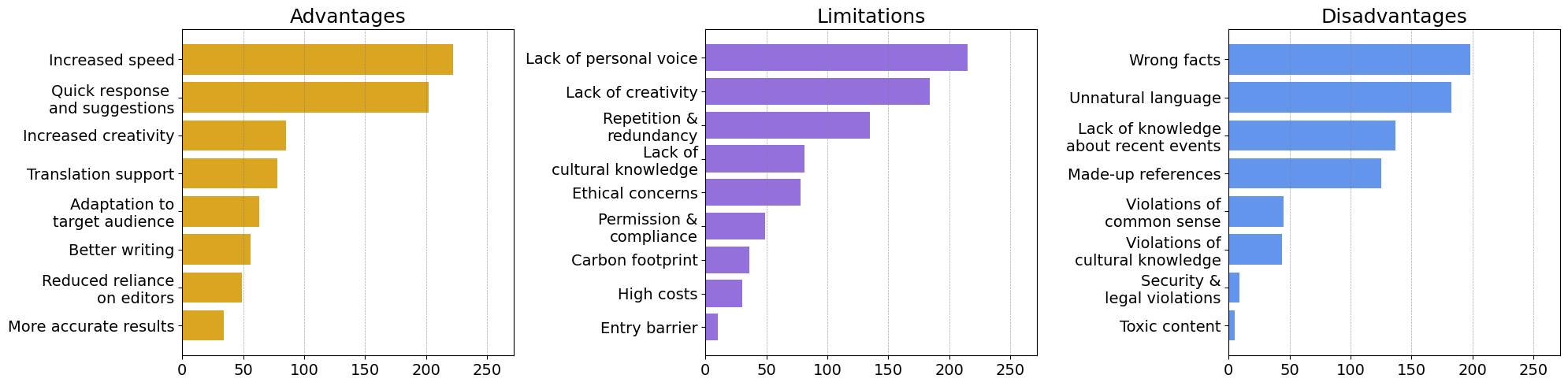}
    \caption{Perceived advantages (left), limitations (center), disadvantages (right) of AI chatbots (N=301).}
    \label{fig:advdisadvlim}
\end{figure*}
\setlength{\textfloatsep}{8pt}

This section summarizes our main findings. Auxiliary analysis regarding some correlations is available in \autoref{subsec:app_results_main_survey}.

\noindent \paragraph{RQ1: Integration of AI chatbots into writing}

\paragraph{Adoption Patterns and User Profiles} AI chatbot adoption is high among professional writers: 73\% of respondents use AI chatbots, with the majority relying on them heavily -- 56\% almost always and 35\% often. This reveals a polarization: writers tend to either avoid AI chatbots entirely or integrate them deeply into their workflow. Factor analysis identified seven user profiles: (1) quality-skeptical users who cross-check AI output with detectors and humanizers; (2) technically-oriented users who optimize content for SEO; (3) efficiency-focused users who prioritize speed and cost reduction; (4) heavy AI chatbot users with paid subscriptions and positive overall impressions; (5) traditional writing task specialists focused on editing, writing, and proofreading; (6) translation specialists for whom translation dominates daily tasks; and (7) ethics-aware users concerned with bias, toxicity, and environmental impact. These profiles can overlap, as individual writers may exhibit characteristics from multiple categories.

Adoption is shaped more by social environment than demographics. Most respondents estimate AI usage among their colleagues at 26–75\%, and perceived colleague usage correlates with both subscription likelihood ($\rho$ = 0.21*) and willingness to pay ($\rho$ = 0.35*), consistent with social influence theory \citep{rogers2003}. Demographics explain little variance in adoption behavior ($R^2$ = 0.02–0.05).

Beyond standard use cases and official capabilities, writers explored interpreting Old Hebrew ~\ref{appendix:quotes}[5], foreign language practice ~\ref{appendix:quotes}[10], emotional support in their native languages ~\ref{appendix:quotes}[6], and assisting older relatives with translations ~\ref{appendix:quotes}[11, 12].

\noindent \paragraph{Tool Preferences and Prompting Strategies} ChatGPT dominates with over 90\% preference, followed at a distance by Gemini, Perplexity, Claude, Copilot, and DeepSeek.\footnote{DeepSeek R1 was launched January 20, 2025, increasing usage between survey stages.} More than 90\% of AI users work primarily in English, while the remaining 10\% use French, German, Arabic, Portuguese, or Russian -- not due to limited English proficiency but reflecting English's dominance as a professional working language.

Writers have developed sophisticated prompting strategies: 75\% improve outputs through iterative prompting and audience specification, while only 7\% use no specific strategies. The interactive task results confirm this: participants engaged in 1 to 18 rounds of interaction, revealing two dominant approaches: specifying tone and audience with targeted revision instructions, and requesting multiple variations to synthesize into a final draft.

\noindent \paragraph{Advantages, Disadvantages, and Impact} Writers broadly value AI for speed, quick responses, enhanced creativity, and translation support (Figure \ref{fig:advdisadvlim}, left), with positive AI assessment correlating with higher valuation ($\rho$ = 0.25**) and subscription likelihood ($\rho$ = 0.23**). Primary disadvantages are unnatural language, factual inaccuracies, and fabricated references (Figure \ref{fig:advdisadvlim}, right); unnatural language concerns correlate with non-payment (${\chi}^2$ = 17.94***), suggesting disadvantages drive non-payment more than advantages drive payment.

Views on creative impact are divided: many describe AI as ``a game changer in the writing world''~\ref{appendix:quotes}[1] and note it has shifted creativity beyond individual authorship~\ref{appendix:quotes}[2]. Others argue that AI still struggles with creativity and originality, especially tone and voice, with \textit{Claude AI} often cited as one of the closer examples~\ref{appendix:quotes}[3]. About half report no writing style change, though frequent usage increases AI's style impact ($\rho$ = 0.21**).

Taken together, RQ1 findings suggest that \textit{AI adoption among professional writers is driven more by professional environment} and creative outlook than by demographic factors.

\paragraph{RQ2: AI Across Languages and Cultures}

\paragraph{Multilingual Usage Patterns}
Nearly half of \textit{multilingual respondents} interact with AI exclusively in English, and among them, 60\% have never attempted non-English interaction. Of those who use AI in other languages, German and French are most popular, followed by Bengali, Zulu, Portuguese, Swahili, Spanish, and Turkish. Most multilingual writers (70\%) write prompts in the target language, while 20\% write in English and translate outputs. The \textit{interactive task} corroborates this behaviorally: a nearly identical share of participants began prompting in the announcement's target language, while 10 of 36 defaulted to English -- fewer than expected given the task's cultural and linguistic specificity, itself a relevant finding on multilingual AI engagement. Even among active multilingual users, English remains the default for professional tasks -- 50\% use AI daily in English for work versus 40\% weekly in other languages for personal purposes.

\paragraph{Performance Gap and Perception Gap}
Writers rate AI performance lower in non-English languages relative to English (rated 10), though the gap varies considerably: Hebrew, Swedish, Portuguese, French, Spanish, German, and Russian average 8 or above, while languages with fewer digital resources score lower (\autoref{tab:language_scores}). Most respondents agree AI struggles with grammar and idioms in non-English languages, though 22\% report comparable performance across languages, particularly in French, German, Spanish, Russian, Arabic, and Swahili. Consistent with this, in 17\% of interactive-task cases (6 of 36 responses, spanning Spanish, Swedish, French, Greek, and German), writers reported no or almost no edits were needed.

The \textit{interactive task} (\S\ref{sec:interactive_task}) provides behavioral evidence of the performance gap directly. Despite expressed concerns, similarity scores between AI drafts and final texts averaged 0.84 -- 58\% required only minor edits, 16\% moderate edits, and 25\% major rewrites. As noted in \S\ref{sec:methodology}, SequenceMatcher is known to inflate similarity for semantically significant edits, so these percentages should be read as a conservative lower bound on human intervention rather than as evidence that AI outputs were largely adequate as-is. The task surfaced concrete, language-specific error patterns that self-report alone cannot capture: style issues in German, word-choice errors in French, overuse of emojis in French and underuse in German, a cultural inaccuracy around Ramadan in Arabic, a content-completeness gap in a Beninese response, and banality in Russian output. Participants also rewrote passages that felt like literal translations from English and corrected cultural inaccuracies in Egyptian Arabic and Swahili, confirming that stylistic and cultural failures require human intervention even when AI produces grammatically acceptable output.

To illustrate the gap between textual footprint and correction stakes, we draw on writers' mandatory self-reports describing what they changed and why. Several writers in the high-similarity (``minor edit'') range described correcting substantive issues despite minimal textual change: a Ukrainian respondent fixed a phrase they described as nonsensical in the original AI output; a Swahili respondent corrected a phrase whose literal meaning did not make sense in context; another respondent corrected a factual error about regional Ramadan food customs; a German respondent corrected a formality-register mismatch inappropriate for the event type. Each involved only a small number of changed words, yet addressed an error the writer themselves identified as important rather than cosmetic -- indicating that similarity scores capture effort but not necessarily the stakes of individual corrections. The full set of self-reported responses from the interactive task is provided in Appendix~\ref{appendix:self_reports}. The causes of variation, whether related to trust, language, or task difficulty, cannot be determined from similarity scores alone.

\paragraph{Cultural Representation}
Writers evaluate cultural representation as distinct from language performance. Among respondents, 13.6\% feel their culture is not represented in AI outputs and 26\% consider it only slightly represented, while 29\% do not use AI for culture-specific content and cannot assess it. Underrepresentation is most reported by writers from African countries (Kenya, Nigeria, Egypt), India, Greece, Cameroon, and Bulgaria; those from the U.S., Germany, U.K., and Canada perceive adequate representation (\autoref{fig:map}).

In free-text responses, writers speaking widely-supported languages more often criticized specific dialect or style gaps, while writers whose languages have only recently gained AI support more often expressed excitement despite acknowledged shortcomings (e.g., \ref{appendix:quotes}[12-15]). We present this as an illustrative qualitative pattern rather than a statistically evaluated group difference; a plausible explanation is differential prior expectations shaped by technology access and cultural visibility, though our data cannot confirm this. Over half of respondents consider linguistic and cultural nuance extremely important, while 9\% deem it unimportant, preferring to modify content themselves. Notably, no correlation exists between these preferences and geographic or language background, suggesting attitudes toward cultural nuance are individually rather than culturally determined.

RQ2 findings reveal that the barriers to equitable multilingual AI use are not only technical. \textit{Performance gaps, perception gaps, and cultural underrepresentation each operate independently}, and writers from non-Western backgrounds face all three simultaneously.

\begin{table*}[h]
    \small
    \centering
    \begin{tabular}{l@{\hskip4pt}c@{\hskip4pt}l@{\hskip10pt}l@{\hskip4pt}c@{\hskip4pt}l@{\hskip10pt}l@{\hskip4pt}c@{\hskip4pt}l@{\hskip10pt}l@{\hskip4pt}c@{\hskip4pt}l}
        \toprule
        \textbf{Language} & \textbf{N} & \textbf{Mean $\pm$ std} & \textbf{Language} & \textbf{N} & \textbf{Mean $\pm$ std} & \textbf{Language} & \textbf{N} & \textbf{Mean $\pm$ std} & \textbf{Language} & \textbf{N} & \textbf{Mean $\pm$ std} \\
        \midrule
        Hebrew     & 2\textsuperscript{\dag}  & 9.5 \tiny{$\pm$0.5}   & German   & 25 & 8.0 \tiny{$\pm$1.47}  & Bengali   & 2\textsuperscript{\dag} & 7.0 \tiny{$\pm$1.0}   & Chinese      & 1\textsuperscript{\dag} & 6.0 \tiny{$\pm$0.0}  \\
        Afrikaans  & 1\textsuperscript{\dag}  & 9.0 \tiny{$\pm$0.0}   & Russian  & 5  & 8.0 \tiny{$\pm$1.10}  & Turkish   & 1\textsuperscript{\dag} & 7.0 \tiny{$\pm$0.0}   & Malay        & 1\textsuperscript{\dag} & 5.0 \tiny{$\pm$0.0}  \\
        Filipino   & 1\textsuperscript{\dag}  & 9.0 \tiny{$\pm$0.0}   & Dutch    & 1\textsuperscript{\dag}  & 8.0 \tiny{$\pm$0.0}   & Ukrainian & 1\textsuperscript{\dag} & 7.0 \tiny{$\pm$0.0}   & Bulgarian    & 2\textsuperscript{\dag} & 4.5 \tiny{$\pm$0.5}  \\
        French     & 24 & 8.5 \tiny{$\pm$1.12}  & Hausa    & 1\textsuperscript{\dag}  & 8.0 \tiny{$\pm$0.0}   & Hindi     & 1\textsuperscript{\dag} & 7.0 \tiny{$\pm$0.0}   & Yoruba       & 1\textsuperscript{\dag} & 4.0 \tiny{$\pm$0.0}  \\
        Portuguese & 2\textsuperscript{\dag}  & 8.5 \tiny{$\pm$1.50}  & Arabic   & 16 & 7.56 \tiny{$\pm$1.12} & Greek     & 3  & 6.33 \tiny{$\pm$0.94} & Bemba        & 1\textsuperscript{\dag} & 3.0 \tiny{$\pm$0.0}  \\
        Swedish    & 2\textsuperscript{\dag}  & 8.5 \tiny{$\pm$0.5}   & Italian  & 2\textsuperscript{\dag}  & 7.5 \tiny{$\pm$1.5}   & Zulu      & 1\textsuperscript{\dag} & 6.0 \tiny{$\pm$0.0}   & Nig. Pidgin  & 1\textsuperscript{\dag} & 2.0 \tiny{$\pm$0.0}  \\
        Spanish    & 11 & 8.27 \tiny{$\pm$1.21} & Swahili  & 8  & 7.25 \tiny{$\pm$2.11} &           &    &                       &              &    &                      \\
        \bottomrule
    \end{tabular}
    \caption{Mean scores per language compared to English (rated as 10). \textsuperscript{\dag} marks languages with $N<3$; these single- or double-respondent estimates (including the several std $=0.0$ entries) should be read as individual data points rather than stable per-language averages. Exact $N$ is reported for every language for full transparency.}
    \label{tab:language_scores}
\end{table*}

\paragraph{RQ3: AI Limitations, Expectations, and Ethics}

\paragraph{Current Limitations}
Over 60\% of writers report loss of personal voice, lack of genuine creativity, repetition, insufficient cultural knowledge, and content ownership concerns as major limitations (\autoref{fig:advdisadvlim}, center). Qualitative responses reveal two deeper concerns: threat to creative voice -- \textit{``The unique voice of human writing gets blurred -- texts start to sound lifeless, correct yet missing a soul''} \ref{appendix:quotes}[4] -- and cognitive autonomy -- \textit{``There is a potential threat that this will significantly diminish creativity and the ability to think and only make us think a certain way — that the AI wants us to think''} \ref{appendix:quotes}[28]. These concerns extend beyond technical shortcomings into creative identity that fact-checking alone will not resolve.

\paragraph{Expectations for Future AI} Writers prioritize trustworthiness over expanded capabilities: fact-checking is the most desired feature (58\%), followed by up-to-date knowledge (35\%), personalization and web search (33\% each), and multimodality (31\%). Notable but less frequent expectations include cultural awareness and support for endangered languages.

\paragraph{Job Security and Authorship}
43\% view AI as a potential threat while believing human expertise remains essential, 23\% perceive minor threat, and 18\% significant threat. Users and non-users do not differ in threat perception, plagiarism concerns, or authorship views — but users hold significantly more positive views of AI progress ($\chi^2$ = 23.79***), suggesting writers adopt AI despite feeling threatened, not because they feel safe.

Authorship attitudes reveal further complexity. Nearly 50\% claim full authorship of AI-assisted work, viewing AI as a tool rather than a co-author, while 18\% support partial user ownership and 19\% remain uncertain. These positions are internally consistent: authorship claimers perceive less job threat ($\rho$ = -0.17**) and disclose AI use less frequently ($\rho$ = -0.19**), while threat-perceivers disclose more ($\rho$ = 0.15**). For disclosure overall, 64\% would voluntarily reveal AI use without formal policies, and half worry about AI-generated content being flagged as plagiarism. Responsible AI development is important to writers: \textit{``Ethical aspects and bias should be prioritized when developing AI chatbots''} \ref{appendix:quotes}[31], with concerns extending to AI misuse in education \ref{appendix:quotes}[33] and insufficient cultural representation \ref{appendix:quotes}[34].

As with adoption, demographics explain minimal variance in ethical attitudes and job security perceptions ($R^2$ = 0.02-0.05). Adding attitude and usage factor scores substantially improves prediction ($R^2$ = 0.14-0.25), confirming that \textit{how writers relate to AI matters far more than who they are demographically}.

\section{Discussion}
Our results align with prior work on AI adoption and authorship, but extend it in two directions: multilingual professional writing and user-side barriers beyond model performance.

\subsection{Widespread Adoption Despite Limitations}
Adoption is driven by peer influence and professional outlook rather than demographics, consistent with social norm accounts of technology uptake \citep{al2025adoption, rogers2003}. Unlike prior work finding limited effects of perceived barriers among researchers \citep{al2025adoption}, concerns about factual errors, unnatural language, and cultural limits persist even among heavy users -- adoption and critical engagement coexist. This contrasts with artistic writing contexts, where AI is framed as ethically unacceptable \citep{lamb2024precaritysolidaritypreliminaryresults, guo2025pen}; our professional writers adopt pragmatically while retaining criticism, suggesting community context and commercial stakes shape resistance more than exposure alone.

Heterogeneous user profiles ranging from quality-skeptical to ethics-engaged argue against interfaces that assume a uniform user model. Large-scale log analysis of real-world interactions identifies similar behavioral diversity \citep{mysore2025prototypical}; our interactive task extends this to multilingual contexts, where writers spend additional revision rounds correcting cultural inaccuracies and translationese rather than refining their original intent, motivating adaptive co-writing interfaces \citep{reza2025co}. The coexistence of heavy AI users and ethics-aware users suggests that addressing concerns about bias, authorship, and transparency may support rather than hinder adoption.

\subsection{The Perception Gap and Sociolinguistic Marginalization}
Among our most consequential findings: some writers never used AI in Swahili, Bulgarian, Turkish, Tagalog, or Georgian because they incorrectly believed those languages were unsupported. This perception gap is distinct from model quality and cannot be closed by better performance, since the barrier is awareness, not capability. Low non-English AI usage reflects three partially independent factors: actual performance limitations, perception gaps, and sociolinguistic conditions that reduce native-language use independently of AI \citep{fishman1991reversing, crystal2002language}. One respondent captures the last: \textit{``Since I left the village, I haven't spoken it because I haven't met anyone who speaks it.''} Model improvements alone are unlikely to address this form of language marginalization.

Cultural underrepresentation compounds these effects. Writers from Africa and India most frequently report missing cultural context, consistent with prior evidence that alignment decreases with cultural distance from Western norms \citep{van2026cultural, biswas2025mind} and remains weaker in lower-resource settings \citep{ying2025disentangling}. Addressing multilingual equity requires both technical and user-facing interventions.

\subsection{Ethical Ambivalence and Authorship}
Professional writers claim full authorship more often than creative writers, who negotiate partial attribution and resist public co-authorship declarations \citep{hwang2024} -- likely because commercial writers anchor identity in editorial control rather than the generative process. \citet{li2024value} show AI assistance reduces perceived ownership experimentally; we complicate this by showing professional users preserve ownership claims through revision gatekeeping, suggesting sustained editorial agency buffers this effect. 

Beyond authorship and ownership, some participants framed AI as extending far beyond a productivity tool, describing it in terms of broader societal and political trajectories. One participant expressed concerns about long-term consequences, warning of \textit{``domination, control, manipulation, and deception''} \ref{appendix:quotes}[26]. While such views are not representative of the majority, they illustrate that a subset of writers situate AI within wider narratives of systemic power and societal change rather than workplace assistance alone.

Threat perception is more moderate than in broader author communities \citep{authorsguild2023}, a gap that community context and precarity help explain \citep{lamb2024precaritysolidaritypreliminaryresults}. Disclosure patterns suggest strategic rather than purely normative behavior: authorship claimers disclose less, whereas threat-perceivers disclose more. Similar tensions around ownership and attribution have been reported in AI co-writing literature \citep{hwang2024, draxler2024ai}. Trust therefore depends not only on model quality but also on how AI contribution is attributed and controlled.

\subsection{Implications for NLP Research}

\textbf{Move beyond automatic metrics.} Professional writers detect stylistic, idiomatic, and cultural failures that benchmarks miss \citep{van2026cultural, ying2025disentangling}. Incorporating them into red-teaming and human evaluation would surface failure modes closer to real deployment conditions.

\textbf{Address the perception gap.} Our findings call for user-facing capability communication -- in-product language indicators and multilingual documentation, not only benchmark scores in technical reports. Closing this gap requires collaboration between NLP researchers and designers to better align system capabilities with how non-English-speaking users perceive AI tools.

\textbf{Include professional writers in development.} Log-based analysis captures behavior \citep{mysore2025prototypical} but not normative judgments about cultural adequacy. Professional writers constitute an underused evaluation population. Unlike crowdworkers or students, they routinely evaluate texts under quality, stylistic, and commercial constraints, making them valuable for human evaluation and deployment-oriented testing.

\section{Conclusion}
Our study of 301 professional writers across 49 countries shows that AI adoption is driven by peer influence and professional environment rather than demographics, while multilingual writers from non-Western backgrounds bear disproportionate corrective labor. Job security concerns are universal and adoption-independent -- daily users remain as worried about displacement as non-users. The perception gap around language support is a communication failure, not a benchmark problem. Writers' expertise in detecting AI failures, particularly cultural ones, is currently invisible to evaluation pipelines; incorporating them into model development is both an equity imperative and a practical necessity.

\section*{Limitations}

\noindent \textbf{Respondent selection.} Participants were recruited from a single crowdsourcing platform, \toloka{}. Although the platform is internationally diverse and includes workers from many countries, the sample may not represent all professional freelance writers. Country composition was not controlled and participation was uneven across regions; therefore, findings should be generalized cautiously to other professional writing settings.

\noindent \textbf{Focus on chat-based AI tools.} We studied general-purpose LLM chatbots (e.g., ChatGPT, Gemini, DeepSeek) rather than specialized writing systems such as grammar checkers, translation tools, or editing assistants. The study therefore reflects broad AI-assisted writing practices and may not generalize to domain-specific tools.

\noindent \textbf{Temporal validity.} Data collection preceded major multilingual updates to recent models. Consequently, language ratings (\autoref{tab:language_scores}) and tool preferences may change as systems improve. However, several findings are less tied to specific model versions, including perception gaps regarding language support, cultural representation concerns, and the relationship between adoption and ethical attitudes, as these reflect user beliefs and professional practices in addition to technical capability.

\noindent \textbf{Behavioral task size.} The interactive writing task involved a smaller subset of participants ($N=36$). While it provided qualitative and behavioral evidence complementing the survey, it was not designed for fine-grained comparisons across languages or user groups.

\section*{Ethical Considerations}

\textbf{Consent and compensation.} The study was conducted through a platform \toloka whose standard moderation process governs research conducted on it. Participation was voluntary and compensated at participants' established hourly rates. All participants were informed of the study's purpose before starting; opening the survey constituted informed consent under the platform's user agreement and privacy notice. Participants had additionally agreed to research reuse of anonymized responses upon platform registration. Relevant excerpts are provided in \autoref{appendix:ethics}.

\textbf{IRB.} Our institution does not have an IRB or equivalent ethics committee. The survey underwent the platform's standard compliance review, and no additional approval was required, as the study was structured as a standard paid task within the platform's existing framework.

\textbf{Anonymization.} All reported data are aggregated and anonymized. No personally identifiable information is released. Qualitative quotes are reported without attribution to individual participants.

\textbf{Use of AI assistance.} We used Grammarly and ChatGPT to correct grammar, spelling, and style in this paper. Specific text segments may therefore be detected as machine-edited or machine-generated.

\section*{Acknowledgements}

We thank Sergei Tilga for his contributions during the initial stages of the project, Artem Sevastopolsky for proofreading, and Toloka for providing access to the platform used in this study.

\bibliography{custom}

@article{spearman1961proof,
  title   = {The Proof and Measurement of Association between Two Things},
  author  = {Spearman, Charles},
  journal = {The American Journal of Psychology},
  volume  = {15},
  number  = {1},
  pages   = {72--101},
  year    = {1904}
}

@article{pearson1900x,
  title={X. On the criterion that a given system of deviations from the probable in the case of a correlated system of variables is such that it can be reasonably supposed to have arisen from random sampling},
  author={Pearson, Karl},
  journal={The London, Edinburgh, and Dublin Philosophical Magazine and Journal of Science},
  volume={50},
  number={302},
  pages={157--175},
  year={1900},
  publisher={Taylor \& Francis}
}

@article{maccallum2002practice,
  title={On the practice of dichotomization of quantitative variables.},
  author={MacCallum, Robert C and Zhang, Shaobo and Preacher, Kristopher J and Rucker, Derek D},
  journal={Psychological methods},
  volume={7},
  number={1},
  pages={19},
  year={2002},
  publisher={American Psychological Association}
}

@article{liao2024llms,
  title={{LLMs as Research Tools: A Large Scale Survey of Researchers' Usage and Perceptions}},
  author={Liao, Zhehui and Antoniak, Maria and Cheong, Inyoung and Cheng, Evie Yu-Yen and Lee, Ai-Heng and Lo, Kyle and Chang, Joseph Chee and Zhang, Amy X},
  journal={arXiv preprint arXiv:2411.05025},
  year={2024}
}

@article{guo2024pen,
  title={{From Pen to Prompt: How Creative Writers Integrate AI into Their Writing Practice}},
  author={Guo, Alicia and Sathyanarayanan, Shreya and Wang, Leijie and Heer, Jeffrey and Zhang, Amy},
  journal={arXiv preprint arXiv:2411.03137},
  year={2024}
}

@article{fecher2023friend,
  title={{Friend or Foe? Exploring the Implications of Large Language Models on the Science System}},
  author={Fecher, Benedikt and Hebing, Marcel and Laufer, Melissa and Pohle, J{\"o}rg and Sofsky, Fabian},
  journal={Ai \& Society},
  pages={1--13},
  year={2023},
  publisher={Springer}
}

@article{morris2023scientists,
  title={{Scientists' Perspectives on the Potential for Generative AI in their Fields}},
  author={Morris, Meredith Ringel},
  journal={arXiv preprint arXiv:2304.01420},
  year={2023}
}

@article{pawar2024survey,
  title={{Survey of Cultural Awareness in Language Models: Text and Beyond}},
  author={Pawar, Siddhesh and Park, Junyeong and Jin, Jiho and Arora, Arnav and Myung, Junho and Yadav, Srishti and Haznitrama, Faiz Ghifari and Song, Inhwa and Oh, Alice and Augenstein, Isabelle},
  journal={arXiv preprint arXiv:2411.00860},
  year={2024}
}

@article{huang2024survey,
  title={A survey on large language models with multilingualism: Recent advances and new frontiers},
  author={Huang, Kaiyu and Mo, Fengran and Zhang, Xinyu and Li, Hongliang and Li, You and Zhang, Yuanchi and Yi, Weijian and Mao, Yulong and Liu, Jinchen and Xu, Yuzhuang and others},
  journal={arXiv preprint arXiv:2405.10936},
  year={2024}
}

@inproceedings{shaib-etal-2024-detection,
    title = "Detection and Measurement of Syntactic Templates in Generated Text",
    author = "Shaib, Chantal  and
      Elazar, Yanai  and
      Li, Junyi Jessy  and
      Wallace, Byron C",
    editor = "Al-Onaizan, Yaser  and
      Bansal, Mohit  and
      Chen, Yun-Nung",
    booktitle = "Proceedings of the 2024 Conference on Empirical Methods in Natural Language Processing",
    month = nov,
    year = "2024",
    address = "Miami, Florida, USA",
    publisher = "Association for Computational Linguistics",
    url = "https://aclanthology.org/2024.emnlp-main.368/",
    doi = "10.18653/v1/2024.emnlp-main.368",
    pages = "6416--6431",
}

@article{myung2024blend,
  title={Blend: A benchmark for llms on everyday knowledge in diverse cultures and languages},
  author={Myung, Junho and Lee, Nayeon and Zhou, Yi and Jin, Jiho and Putri, Rifki and Antypas, Dimosthenis and Borkakoty, Hsuvas and Kim, Eunsu and Perez-Almendros, Carla and Ayele, Abinew Ali and others},
  journal={Advances in Neural Information Processing Systems},
  volume={37},
  pages={78104--78146},
  year={2024}
}

@article{biswas2025mind,
  title={Mind the Gap! Choice Independence in Using Multilingual LLMs for Persuasive Co-Writing Tasks in Different Languages},
  author={Biswas, Shreyan and Erlei, Alexander and Gadiraju, Ujwal},
  journal={arXiv preprint arXiv:2502.09532},
  year={2025}
}

@inproceedings{blaschke-etal-2024-dialect,
    title = "What Do Dialect Speakers Want? A Survey of Attitudes Towards Language Technology for {G}erman Dialects",
    author = "Blaschke, Verena  and
      Purschke, Christoph  and
      Schuetze, Hinrich  and
      Plank, Barbara",
    editor = "Ku, Lun-Wei  and
      Martins, Andre  and
      Srikumar, Vivek",
    booktitle = "Proceedings of the 62nd Annual Meeting of the Association for Computational Linguistics (Volume 2: Short Papers)",
    month = aug,
    year = "2024",
    address = "Bangkok, Thailand",
    publisher = "Association for Computational Linguistics",
    url = "https://aclanthology.org/2024.acl-short.74/",
    doi = "10.18653/v1/2024.acl-short.74",
    pages = "823--841",
}

@misc{stackoverflow2024,
  author       = {{Stack Overflow}},
  title        = {2024 Stack Overflow Developer Survey},
  year         = {2024},
  url          = {https://survey.stackoverflow.co/2024},
  note         = {Accessed: 2025-03-19}
}

@article{draxler2024ai,
  title={The AI ghostwriter effect: When users do not perceive ownership of AI-generated text but self-declare as authors},
  author={Draxler, Fiona and Werner, Anna and Lehmann, Florian and Hoppe, Matthias and Schmidt, Albrecht and Buschek, Daniel and Welsch, Robin},
  journal={ACM Transactions on Computer-Human Interaction},
  volume={31},
  number={2},
  pages={1--40},
  year={2024},
  publisher={ACM New York, NY}
}

@incollection{kellogg2018,
  author    = {Kellogg, Ronald T.},
  title     = {Professional Writing Expertise},
  booktitle = {The Cambridge Handbook of Expertise and Expert Performance},
  editor    = {Ericsson, K. Anders and Hoffman, Robert R. and Kozbelt, Aaron and Williams, A. Mark},
  series    = {Cambridge Handbooks in Psychology},
  publisher = {Cambridge University Press},
  address   = {Cambridge},
  year      = {2018},
  pages     = {413--430}
}

@misc{chatgpt,
  author    = {OpenAI},
  title     = {ChatGPT},
  year      = {2025},
  url       = {https://chat.openai.com},
  note      = {Accessed: 2025-03-28}
}

@article{scikit-learn,
  title={Scikit-learn: Machine Learning in {P}ython},
  author={Pedregosa, F. and Varoquaux, G. and Gramfort, A. and Michel, V.
          and Thirion, B. and Grisel, O. and Blondel, M. and Prettenhofer, P.
          and Weiss, R. and Dubourg, V. and Vanderplas, J. and Passos, A. and
          Cournapeau, D. and Brucher, M. and Perrot, M. and Duchesnay, E.},
  journal={Journal of Machine Learning Research},
  volume={12},
  pages={2825--2830},
  year={2011}
}

@book{rogers2003,
  address = {New York, NY [u.a.]},
  author = {Rogers, Everett M.},
  edition = {5th},
  isbn = {0-7432-2209-1, 978-0-7432-2209-9},
  language = {eng},
  month = {08},
  pages = 576,
  publisher = {Free Press},
  title = {Diffusion of innovations},
  year = 2003
}

@book{hair2013multivariate,
  title={Multivariate data analysis: Pearson new international edition PDF eBook},
  author={Hair, Joseph F and Black, William C and Babin, Barry J and Anderson, Rolph E},
  year={2013},
  publisher={Pearson Higher Ed}
}

@article{al2025adoption,
  title={Adoption of AI writing tools among academic researchers: A Theory of Reasoned Action approach},
  author={Al-Bukhrani, Mohammed A and Alrefaee, Yasser Mohammed Hamid and Tawfik, Mohammed},
  journal={PloS one},
  volume={20},
  number={1},
  pages={e0313837},
  year={2025},
  publisher={Public Library of Science San Francisco, CA USA}
}

@inproceedings{guo2025pen,
  title={From pen to prompt: How creative writers integrate AI into their writing practice},
  author={Guo, Alicia and Sathyanarayanan, Shreya and Wang, Leijie and Heer, Jeffrey and Zhang, Amy X},
  booktitle={Proceedings of the 2025 Conference on Creativity and Cognition},
  pages={527--545},
  year={2025}
}

@article{reza2025co,
  title={Co-writing with AI, on human terms: Aligning research with user demands across the writing process},
  author={Reza, Mohi and Thomas-Mitchell, Jeb and Dushniku, Peter and Laundry, Nathan and Williams, Joseph Jay and Kuzminykh, Anastasia},
  journal={Proceedings of the ACM on Human-Computer Interaction},
  volume={9},
  number={7},
  pages={1--37},
  year={2025},
  publisher={ACM New York, NY, USA}
}

@book{fishman1991reversing,
  title={Reversing language shift: Theoretical and empirical foundations of assistance to threatened languages},
  author={Fishman, Joshua A},
  volume={76},
  year={1991},
  publisher={Multilingual matters}
}

@book{crystal2002language,
  title={Language death},
  author={Crystal, David},
  year={2002},
  publisher={Cambridge university press}
}

@inproceedings{bender2021dangers,
  title={On the dangers of stochastic parrots: Can language models be too big?},
  author={Bender, Emily M and Gebru, Timnit and McMillan-Major, Angelina and Shmitchell, Shmargaret},
  booktitle={Proceedings of the 2021 ACM conference on fairness, accountability, and transparency},
  pages={610--623},
  year={2021}
}

@techreport{authorsguild2023,
  author       = {{The Authors Guild}},
  title        = {Authors Guild Annual Report for 2023},
  institution  = {The Authors Guild},
  year         = {2023},
  url          = {https://authorsguild.org/app/uploads/2024/03/2023-Authors-Guild-Annual-Report.pdf},
  note         = {Accessed: 2026-05-25}
}

@article{braun2006,
  title={Using thematic analysis in psychology},
  author={Braun, Virginia and Clarke, Victoria},
  journal={Qualitative research in psychology},
  volume={3},
  number={2},
  pages={77--101},
  year={2006},
  publisher={Taylor \& Francis}
}

@inproceedings{mysore2025prototypical,
  title={Prototypical human-AI collaboration behaviors from LLM-Assisted writing in the wild},
  author={Mysore, Sheshera and Das, Debarati and Cao, Hancheng and Sarrafzadeh, Bahareh},
  booktitle={Proceedings of the 2025 Conference on Empirical Methods in Natural Language Processing},
  pages={16830--16857},
  year={2025}
}

@misc{lamb2024precaritysolidaritypreliminaryresults,
      title={Precarity and Solidarity: Preliminary results on a study of queer and disabled fiction writers' experiences with generative AI}, 
      author={C. E. Lamb and D. G. Brown and M. R. Grossman},
      year={2024},
      eprint={2412.04575},
      archivePrefix={arXiv},
      primaryClass={cs.CY},
      url={https://arxiv.org/abs/2412.04575}, 
}

@misc{li2024value,
      title={The Value, Benefits, and Concerns of Generative AI-Powered Assistance in Writing}, 
      author={Zhuoyan Li and Chen Liang and Jing Peng and Ming Yin},
      year={2024},
      eprint={2403.12004},
      archivePrefix={arXiv},
      primaryClass={cs.HC},
      url={https://arxiv.org/abs/2403.12004}, 
}

@article{hu2025negotiating,
  title={Negotiating identity in the age of ChatGPT: non-native English researchers’ experiences with AI-assisted academic writing},
  author={Hu, Hengzhi and Zhou, Qing and Hashim, Harwati},
  journal={Humanities and Social Sciences Communications},
  volume={12},
  number={1},
  pages={1--11},
  year={2025},
  publisher={Palgrave}
}

@inproceedings{ying2025disentangling,
  title={Disentangling language and culture for evaluating multilingual large language models},
  author={Ying, Jiahao and Tang, Wei and Zhao, Yiran and Cao, Yixin and Rong, Yu and Zhang, Wenxuan},
  booktitle={Proceedings of the 63rd Annual Meeting of the Association for Computational Linguistics (Volume 1: Long Papers)},
  pages={22230--22251},
  year={2025}
}

@article{van2026cultural,
  title={Cultural Authenticity: Comparing LLM Cultural Representations to Native Human Expectations},
  author={van Liemt, Erin MacMurray and Davani, Aida and Kumbale, Sinchana and Dixit, Neha and Dev, Sunipa},
  journal={arXiv preprint arXiv:2604.03493},
  year={2026}
}

@inproceedings{guo2025care,
  title={CARE: Multilingual human preference learning for cultural awareness},
  author={Guo, Geyang and Naous, Tarek and Wakaki, Hiromi and Nishimura, Yukiko and Mitsufuji, Yuki and Ritter, Alan and Xu, Wei},
  booktitle={Proceedings of the 2025 Conference on Empirical Methods in Natural Language Processing},
  pages={32854--32883},
  year={2025}
}

@misc{hwang2024,
      title={``{I}t Was 80\% Me, 20\% {AI}'': Seeking Authenticity in Co-Writing with Large Language Models}, 
      author={Angel Hsing-Chi Hwang and Q. Vera Liao and Su Lin Blodgett and Alexandra Olteanu and Adam Trischler},
      year={2024},
      eprint={2411.13032},
      archivePrefix={arXiv},
      primaryClass={cs.HC},
      url={https://arxiv.org/abs/2411.13032}, 
}
\appendix

\clearpage
\newpage

\section{Survey Questions} \label{appendix:survey}
\newlist{questions}{enumerate}{1}
\setlist[questions,1]{label=\arabic*., left=0pt, itemsep=1em}

\newcommand{\question}{\item}

\newlist{checkboxes}{itemize}{1}
\setlist[checkboxes]{label=$\square$, left=1.5em, itemsep=0.5em}

\newlist{choices}{itemize}{1}
\setlist[choices]{label=$\bigcirc$, left=1.5em, itemsep=0.5em}

\newcommand{\choice}{\item}

\renewcommand{\arraystretch}{1.5}

\subsection*{Introduction} \label{sec:app_survey_sec1_intro}

This section reproduces the full questionnaire administered to survey participants ($N=301$), including all question wording and response formats.

\begin{questions}

\question Payment processing question 1

\question Payment processing question 2

\question What is your country of birth?  

\textit{Mark only one option}
\begin{checkboxes}
    \choice Choices omitted for brevity
\end{checkboxes}

\question What is your country of residence?  

\textit{Mark only one option}
\begin{checkboxes}
    \choice Choices omitted for brevity
\end{checkboxes}

\question What is your age?  

\textit{Mark only one option}
\begin{checkboxes}
    \choice 18-25 years 
    \choice 26-35 years 
    \choice 36-45 years 
    \choice 46-55 years 
    \choice 56-65 years 
    \choice 66+ years
\end{checkboxes}

\question What is your highest attained level of education? 

\textit{Mark only one option}
\begin{checkboxes}
    \choice High school 
    \choice Undergraduate degree 
    \choice Postgraduate degree 
    \choice Master's degree 
    \choice Doctoral degree
    \choice None of the above
\end{checkboxes}

\question Is your education connected with professional writing?

\textit{Mark only one option}
\begin{checkboxes}
    \choice Yes
    \choice No
\end{checkboxes}

\question What is your gender?  

\textit{Mark only one option}
\begin{checkboxes}
    \choice Male
    \choice Female
    \choice Prefer not to say 
    \choice Other
\end{checkboxes}

\question What is your preferred language for thinking and reasoning?  \textit{Mark only one option}
\begin{checkboxes}
    \choice English
    \choice Other language
\end{checkboxes}

\question How many years of work experience related to writing do you have?  

\textit{Mark only one option}
\begin{checkboxes}
    \choice Less than 1 year 
    \choice 1-3 years
    \choice 3-5 years
    \choice 6-10 years
    \choice 10+ years
\end{checkboxes}

\question Do you have another part-time or full-time activity (job or academic education) that doesn't involve writing?  

\textit{Mark only one option}
\begin{checkboxes}
    \choice Yes
    \choice No
\end{checkboxes}

\question What is your writing specialization? 

\textit{Mark all options that apply}
\begin{checkboxes}
    \choice Journalism
    \choice Marketing and advertising 
    \choice Technical writing
    \choice PR and communications 
    \choice Academic and educational writing 
    \choice Freelance and online content 
    \choice Corporate and legal writing 
    \choice Translation
    \choice Other: \makebox[1in]{\hrulefill}
\end{checkboxes}

\question What industry have you worked in? 

\textit{Mark all options that apply}
\begin{checkboxes}
    \choice Education
    \choice Media \& communication 
    \choice Healthcare
    \choice Finance
    \choice Technology Manufacturing
    \choice Legal \& law
    \choice Retail
    \choice Tourism
    \choice Other: \makebox[1in]{\hrulefill}
\end{checkboxes}

\question What are your typical daily writing tasks? 

\textit{Mark all options that apply}
\begin{checkboxes}
    \choice Write
    \choice Edit
    \choice Translate
    \choice Proofread
    \choice Research for new content
    \choice Optimize for SEO
    \choice Analyze content performance / Quality assurance 
    \choice Communicate with management and teams
    \choice Other: \makebox[1in]{\hrulefill}
\end{checkboxes}

\question How often do you use English in professional contexts (e.g. for writing)?  

\textit{Mark only one option}
\begin{checkboxes}
    \choice Daily
    \choice Several times a week
    \choice Once a week
    \choice A few times a month 
    \choice Rarely or never
\end{checkboxes}

\question How do you assess your proficiency in English for writing skills?

\textit{Mark only one option}
\begin{checkboxes}
    \choice Excellent 
    \choice Good 
    \choice Fair 
    \choice Poor 
    \choice Beginner
\end{checkboxes}

\question Which language do you primarily use when interacting with AI chatbots?

\textit{Mark only one option}
\begin{checkboxes}
    \choice English
    \choice Other language
\end{checkboxes}

\question In your everyday tasks, do you use AI chatbots like ChatGPT? 

\textit{Mark only one option}
\begin{checkboxes}
    \choice Yes $\rightarrow$ Skip to question 19
    \choice No $\rightarrow$ Skip to question 25
\end{checkboxes}

\subsection*{Using AI chatbots in writing tasks} \label{sec:app_survey_sec1_using_AI}

\question Which AI chatbots do you use, and how often? 

\textit{(Mark only one option per row.)}

\begin{center}
\resizebox{0.85\linewidth}{!}{%
\begin{tabular}{l*{4}{>{\centering\arraybackslash}p{2cm}}}
\toprule
 & \textbf{Never} & \textbf{Rarely} & \textbf{Often} & \textbf{Almost always} \\
\midrule
\textbf{ChatGPT}    & $\bigcirc$ & $\bigcirc$ & $\bigcirc$ & $\bigcirc$ \\
\textbf{Perplexity} & $\bigcirc$ & $\bigcirc$ & $\bigcirc$ & $\bigcirc$ \\
\textbf{Gemini}     & $\bigcirc$ & $\bigcirc$ & $\bigcirc$ & $\bigcirc$ \\
\textbf{Copilot}    & $\bigcirc$ & $\bigcirc$ & $\bigcirc$ & $\bigcirc$ \\
\bottomrule
\end{tabular}
}
\end{center}

\question If anything other (e. g. self-hosted LLMs), please specify:

\makebox[1in]{\hrulefill}

\question How often do you use AI chatbots for each of the following tasks?

\textit{(Mark only one option per row)}

\begin{center}
\resizebox{0.85\linewidth}{!}{%
\begin{tabular}{p{4cm}*{4}{>{\centering\arraybackslash}p{2cm}}}
\toprule
 & \textbf{Never} & \textbf{Rarely} & \textbf{Often} & \textbf{Almost always} \\
\midrule
\textbf{Writing content from scratch}    & $\bigcirc$ & $\bigcirc$ & $\bigcirc$ & $\bigcirc$ \\
\textbf{Expanding short texts into longer, more detailed ones} & $\bigcirc$ & $\bigcirc$ & $\bigcirc$ & $\bigcirc$ \\
\textbf{Improving the quality of a text (e.g., style, grammar, clarity)}     & $\bigcirc$ & $\bigcirc$ & $\bigcirc$ & $\bigcirc$ \\
\textbf{Summarizing texts}    & $\bigcirc$ & $\bigcirc$ & $\bigcirc$ & $\bigcirc$ \\
\textbf{Analyzing a text (e.g., identifying themes, extracting key points)}    & $\bigcirc$ & $\bigcirc$ & $\bigcirc$ & $\bigcirc$ \\
\textbf{Translating text}    & $\bigcirc$ & $\bigcirc$ & $\bigcirc$ & $\bigcirc$ \\
\textbf{Answering questions based on a text}    & $\bigcirc$ & $\bigcirc$ & $\bigcirc$ & $\bigcirc$ \\
\textbf{Brainstorming ideas}    & $\bigcirc$ & $\bigcirc$ & $\bigcirc$ & $\bigcirc$ \\
\textbf{Retrieving relevant web pages or sources}    & $\bigcirc$ & $\bigcirc$ & $\bigcirc$ & $\bigcirc$ \\
\textbf{Creating outlines or structured plans}    & $\bigcirc$ & $\bigcirc$ & $\bigcirc$ & $\bigcirc$ \\
\textbf{Paraphrasing or rewording content}    & $\bigcirc$ & $\bigcirc$ & $\bigcirc$ & $\bigcirc$ \\
\textbf{Conducting language-specific tasks (checking tone or cultural nuances)}    & $\bigcirc$ & $\bigcirc$ & $\bigcirc$ & $\bigcirc$ \\
\textbf{Formatting content (e.g. creatinh references)}    & $\bigcirc$ & $\bigcirc$ & $\bigcirc$ & $\bigcirc$ \\
\bottomrule
\end{tabular}
}
\end{center}

\question Do you use any specific tricks or strategies to improve the quality of answers from AI chatbots?

\textit{(Mark only one option)}
\begin{checkboxes}
    \choice I focus on clarifying requests or specifying the target audience
    \choice I use other specific strategies to improve responses
    \choice No, I don’t use any strategies
\end{checkboxes}

\question Do you pay for using AI chatbots?

\textit{(Mark only one option)}
\begin{checkboxes}
    \choice Yes, I have a subscription
    \choice No, but I had a subscription in the past
    \choice No, I don't have a subscription
    \choice No, but I might consider getting a subscription in the near future
    
\end{checkboxes}

\question What is the maximum amount you would be willing to spend per month on your favorite AI chatbot, expressed in the equivalent number of coffee cups (based on the average cost of one cup of coffee you typically buy)?

\textit{(Mark only one option)}
\begin{checkboxes}
    \choice Up to 5 coffee cups per month
    \choice 6 to 10 coffee cups per month
    \choice 11 to 20 coffee cups per month
    \choice 21 to 30 coffee cups per month 
    \choice More than 30 coffee cups per month
    \choice I would not pay for an AI chat bots
\end{checkboxes}

\question What are the main advantages of using AI chatbots for professional writing?

\textit{(Mark up to three options)}
\begin{checkboxes}
    \choice Increased speed
    \choice Increased creativity
    \choice Better writing
    \choice More accurate results Translation support
    \choice Adaptation to target audience 
    \choice Quick response and suggestions
    \choice Reduced reliance on editors
    \choice Other: \makebox[1in]{\hrulefill}
\end{checkboxes}

\question What percentage of your writing colleagues do you think use AI chatbots in their work, assuming it is not forbidden?

\textit{(Mark only one option)}
\begin{checkboxes}
    \choice Less than 10\% 
    \choice 10-25\% 
    \choice 26-50\%
    \choice 51-75\%
    \choice More than 75\%
    \choice I don't know
\end{checkboxes}

\question Do you feel AI chatbots introduce things from the lists? 

\textit{(Mark up to three options)}
\begin{checkboxes}
    \choice Wrong facts
    \choice Violations of common sense and world knowledge
    \choice Toxic content
    \choice Unnatural language
    \choice Violations of culture-specific knowledge
    \choice Lack of knowledge about recent events
    \choice Providing infomation that violaties security and legal standarts 
    \choice Unexisting and made-up references that back up responses
    \choice Other: \makebox[1in]{\hrulefill}
\end{checkboxes}

\question Do you think AI chatbots influenced your writing style or voice?

\textit{(Mark only one option)}
\begin{checkboxes}
    \choice Not at all – My writing style and voice remain unchanged
    \choice Slightly – I’ve noticed minor changes, but they are not significant 
    \choice Moderately – AI chatbots have had a noticeable impact on my writing style 
    \choice Completely – My writing style or voice has been transformed by AI chatbots
    \choice I’m not sure – I haven’t observed any clear changes
\end{checkboxes}

\question Are you able to tell if a text is AI-generated?

\textit{(Mark only one option)}
\begin{checkboxes}
    \choice Yes, in most cases by my impression
    \choice Yes, about 50\% of the time by my impression 
    \choice Yes, but quite rarely
    \choice Not at all or almost never
\end{checkboxes}

\question Do you agree that professional writers can recognize AI-generated texts better than non-experts?

\textit{(Mark only one option)}
\begin{checkboxes}
    \choice Strongly agree
    \choice Agree
    \choice Disagree
    \choice Strongly disagree 
    \choice Don't know
\end{checkboxes}

\subsection*{Using AI chatbots in non-English languages} \label{sec:app_survey_sec1_using_AI_non_English}

\question What languages other than English can you speak fluently or natively? Please list them, separated by commas:

\makebox[1in]{\hrulefill}

\question If there are any details on dialects you speak, please specify:

\makebox[1in]{\hrulefill}

\question Do you use AI chatbots in any language other than English?

\textit{(Mark only one option)}
\begin{checkboxes}
    \choice Yes $\rightarrow$ Skip to question 34
    \choice Yes $\rightarrow$ Skip to question 44
\end{checkboxes}

\subsection*{Using AI chatbots in non-English languages, part 2}

Let's call the language in which you use AI chatbots most (excluding English) the Language*.

\question What is your Language*?

\makebox[1in]{\hrulefill}

\question How do you assess your proficiency in the Language* for writing skills?

\begin{center}
\resizebox{0.85\linewidth}{!}{%
\begin{tabular}{l*{5}{>{\centering\arraybackslash}p{2cm}}}
\toprule
\textbf{1 Beginner} & \textbf{2} & \textbf{3} & \textbf{4} & \textbf{5 Native} \\
\midrule
$\bigcirc$    & $\bigcirc$ & $\bigcirc$ & $\bigcirc$ & $\bigcirc$ \\

\bottomrule
\end{tabular}
}
\end{center}

\question How often do you use this Language* in professional contexts (e.g., for writing)?

\textit{(Mark only one option)}
\begin{checkboxes}
    \choice Daily
    \choice Several times a week
    \choice Once a week
    \choice A few times a month 
    \choice Rarely or never
\end{checkboxes}

\question How often do you use this Language* for personal needs?

\textit{(Mark only one option)}
\begin{checkboxes}
    \choice Daily
    \choice Several times a week
    \choice Once a week
    \choice A few times a month 
    \choice Rarely or never
\end{checkboxes}

\question Check all tasks you use AI chatbots for in the Language*

\textit{Mark all options that apply}
\begin{checkboxes}
    \choice Writing content from scratch
    \choice Expanding short texts into longer, more detailed ones
    \choice Improving the quality of a text (e.g., style, grammar, clarity)
    \choice Summarizing texts
    \choice Analyzing a text (e.g., identifying themes, extracting key points)
    \choice Translating text
    \choice Answering questions based on a text
    \choice Brainstorming ideas
    \choice Retrieving relevant web pages or sources
    \choice Creating outlines or structured plans
    \choice Paraphrasing or rewording content
    \choice Conducting language-specific tasks (e.g., checking tone or cultural nuances)     \choice Formatting content (e. g. creating references)
\end{checkboxes}

\question If your favorite AI chatbot performs at 10 out of 10 in English, what would its score be in Language*?

\textit{(Mark only one option)}
\begin{center}
\resizebox{0.85\linewidth}{!}{%
\begin{tabular}{l*{10}{>{\centering\arraybackslash}p{1cm}}}
\toprule
\textbf{1} & \textbf{2} & \textbf{3} & \textbf{4} & \textbf{5} & \textbf{6} & \textbf{7} & \textbf{8} & \textbf{9} & \textbf{10}\\
\midrule
$\bigcirc$    & $\bigcirc$ & $\bigcirc$ & $\bigcirc$ & $\bigcirc$ & $\bigcirc$    & $\bigcirc$ & $\bigcirc$ & $\bigcirc$ & $\bigcirc$\\

\bottomrule
\end{tabular}
}
\end{center}

\question How does the use of AI chatbots differ between English and the Language*?

\textit{(Mark up to three options)}
\begin{checkboxes}
    \choice There are fewer tools or platforms available for working in Language*
    \choice AI performs better in English, with more accurate and relevant outputs
    \choice AI tends to struggle with grammar, vocabulary, or idiomatic expressions in the Language*
    \choice AI works better in English for certain tasks (e.g., creative writing, summarization, etc.)
    \choice AI is more culturally accurate or contextually appropriate in English than in the Language*
    \choice The experience is similar for both English and the Language*, with no significant difference
    \choice Other: \makebox[1in]{\hrulefill}
\end{checkboxes}

\question In which way do you mostly write prompts when working in your Language*?

\textit{(Mark only one option)}
\begin{checkboxes}
    \choice Write prompts in English, then translate the generated output into the Language*
    \choice Write prompts directly in the Language* using the common script for that language
    \choice Write prompts phonetically in the Language* using English letters
    \choice Other: \makebox[1in]{\hrulefill}
\end{checkboxes}

\question How do you get quality answers from AI chatbots in your Language*?

\textit{(Mark only one option)}
\begin{checkboxes}
    \choice In the same way I do for English.
    \choice In nearly the same way I do for English, but I should manually edit the response more.
    \choice Not in the same way I do for English
\end{checkboxes}

\question Feel free to share your opinions regarding AI chatbots in your Language*. For instance, if you have interesting use cases (e.g., you use AI in the Language* at the translation request of your older relatives) 

$\rightarrow$ Skip to question 46

\makebox[1in]{\hrulefill}

\question Why don't you use AI chatbots in languages other than English?

\textit{(Mark only one option)}
\begin{checkboxes}
    \choice In the same way I do for English.
    \choice In nearly the same way I do for English, but I should manually edit the response more.
    \choice Not in the same way I do for English
\end{checkboxes}

\question Have you ever tried to use AI chatbots in any of your languages?

\textit{(Mark only one option)}
\begin{checkboxes}
    \choice Yes, and it worked perfectly 
    \choice Yes, and it worked sufficiently
    \choice Yes, and it poorly
    \choice No
\end{checkboxes}

\subsection*{Cultural nuances in the use of AI chatbots} \label{sec:app_survey_sec1_cultural}

\question Do you feel your culture is adequately represented in AI chatbots, regardless of the language used?

\textit{(Mark only one option)}
\begin{checkboxes}
    \choice Not at all represented – There are many errors in culture-specific content and linguistic nuances
    \choice Slightly represented
    \choice Well represented
    \choice Very well represented – Usually there are no errors in culture-specific content and
    \choice I do not know, I don't use AI for culture-specific content
\end{checkboxes}

\question How important is it for you personally that AI chatbots respect linguistic and cultural nuances?

\textit{(Mark only one option)}
\begin{checkboxes}
    \choice Not at all important – I can always modify the content
    \choice Moderately important
    \choice Extremely important – I would not use a tool that makes errors in culturally specific content or fails to respect linguistic nuances
\end{checkboxes}

\question How do you mostly deal with AI responses that don’t match your language or culture well? Choose the best applying option.

\textit{(Mark only one option)}
\begin{checkboxes}
    \choice Manually change the output to better fit the language or culture
    \choice Use specific prompts to make the responses more accurate
    \choice Lack of cultural understanding doesn’t significantly impact my work
    \choice Use AI chatbots mainly for ideas or structure and handle cultural nuances manually
    \choice Avoid using AI chatbots for tasks that need cultural understanding
\end{checkboxes}

\question What other tools do you use for professional writing besides AI chatbots?

\textit{Mark all options that apply}
\begin{checkboxes}
    \choice Spellcheckers
    \choice Voice input (dictation)
    \choice Voice output (text-to-speech) SEO tools
    \choice Translation tools
    \choice Plagiarism detectors Paraphrasing tools
    \choice Dictionaries and corpora
    \choice AI detectors
    \choice Humanizers of AI-generated texts
\end{checkboxes}

\question What are the limitations of using AI chatbots for professional writing?

\textit{(Mark up to three options)}
\begin{checkboxes}
    \choice Lack of genuine creativity: AI generates clichés or predictable texts
    \choice Loss of personal voice: Generated texts seem less authentic and dilute the writer’s unique voice
    \choice Lack of cultural knowledge: AI isn’t able to generate content appropriate for specific audiences or cultures
    \choice Repetition and redundancy: Generated content reads as repetitive, blunt, and redundant
    \choice Costs: AI chatbots are expensive to use
    \choice Carbon footprint: AI chatbots generate a significant amount of CO\textsubscript{2}
    \choice Entry barrier: It takes a lot of time to learn how to use AI chatbots and prompt them effectively
    \choice Ethical concerns: One can’t be sure they own the work written with the help of an AI chatbot
    \choice Compliance with policies: One can't be sure they won't be penalized for using AI chatbots
    \choice Other: \makebox[1in]{\hrulefill}
\end{checkboxes}

\question What are your expectations for the next generation of AI chatbots? Please choose up to 3 options from the list.

\textit{(Mark up to three options)}
\begin{checkboxes}
    \choice Speech processing: Speech recognition and synthesis for voice input and output
    \choice Multimodality: Processing texts together with images or audio
    \choice Fact-checking: The ability to cite reliable sources for factual data
    \choice Web search: The ability to search on the internet and reference relevant sites
    \choice Domain-specific knowledge: Understanding of university-level topics
    \choice Culture awareness: Understanding of diverse cultural contexts
    \choice Personalization: The ability to remember and adapt to user preferences and requirements
    \choice Safety: Improved bias and toxicity mitigation with prevention of harmful idea perpetuation
    \choice Efficiency: Reduced costs and increased speed
    \choice Up-to-date knowledge: Understanding of recent events
    \choice User experience: Improved user interface and enhanced usability
\end{checkboxes}

\question Do you consider yourself open to adopting new technologies in general? 

\begin{center}
\resizebox{0.85\linewidth}{!}{%
\begin{tabular}{l*{5}{>{\centering\arraybackslash}p{2cm}}}
\toprule
\textbf{1 Not open} & \textbf{2} & \textbf{3} & \textbf{4} & \textbf{5 Open} \\
\midrule
$\bigcirc$    & $\bigcirc$ & $\bigcirc$ & $\bigcirc$ & $\bigcirc$ \\

\bottomrule
\end{tabular}
}
\end{center}

\question How did you learn to use AI chatbots?

\textit{Mark all options that apply}
\begin{checkboxes}
    \choice Trial and error
    \choice Self-taught through online tutorials and resources Formal training or courses
    \choice Help from colleagues or peers
    \choice Official documentation and user manuals Community forums and discussion groups
\end{checkboxes}

\question Which phrase best describes your first impression of AI chatbots?

\textit{(Mark only one option)}
\begin{checkboxes}
    \choice I was skeptical and had low expectations
    \choice I found them confusing and difficult to use initially
    \choice I was neutral and didn't have strong feelings either way. I was disappointed by the initial results
    \choice I had a decent experience but wasn't overly impressed. I was surprised by how well they performed
\end{checkboxes}

\question Which phrase best describes your current impression of AI chatbots?

\textit{(Mark only one option)}
\begin{checkboxes}
    \choice I remain skeptical and have low expectations
    \choice I find them confusing and difficult to use
    \choice I am neutral and don't have strong feelings either way
    \choice I am disappointed by their current results
    \choice I have a decent experience but am not overly impressed. I am surprised by how well they continue to perform
\end{checkboxes}

\question Are you interested in the topic of AI in general? To what extent?

\textit{(Mark only one option)}
\begin{checkboxes}
    \choice Very interested: I actively seek out information, follow specialized AI channels
    \choice Moderately interested: I follow AI news and trends but don't actively seek out detailed information
    \choice Somewhat interested: I have a general interest in AI and read news when I come across it
    \choice Slightly interested: I have a passing interest in AI but don't engage with it often
    \choice Not interested: I do not think I'm interested in the topic and do not actively follow AI news or developments
\end{checkboxes}

\question Check the terms you are familiar with in the context of AI chatbots:

\textit{Mark all options that apply}
\begin{checkboxes}
    \choice Transformer models 
    \choice Large Language Models 
    \choice AI hallucinations 
    \choice Multihead attention 
    \choice Reinforcement learning 
    \choice Loss function
    \choice GPU
    \choice Gradient
    \choice Supervised learning 
    \choice Unsupervised learning 
    \choice Pre-trained models 
    \choice Model fine-tuning 
    \choice Training data
\end{checkboxes}

\subsection*{AI impact and ethical issues} \label{sec:app_survey_sec1_AI_impact}

\question To what extent do you view AI chatbots as a potential threat to your profession?

\textit{(Mark only one option)}
\begin{checkboxes}
    \choice Significant threat: I view AI as a significant threat to my profession and job security
    \choice Potential threat: I see AI as a potential threat, but I believe there will always be a need for human expertise in my field
    \choice Minor threat: I don't view AI as a major threat, but I recognize it could change the way we work
    \choice Neutral: I am neutral and unsure whether AI will have a major impact on my profession
    \choice No threat: I don’t view AI as a threat at all; it is just another tool for productivity
\end{checkboxes}

\question What do you think about the progress in AI in general?

\textit{(Mark only one option)}
\begin{checkboxes}
    \choice Very positive: I believe AI is making significant strides and will have a transformative impact on society
    \choice Positive: I think AI is progressing well and will bring many benefits, but there are also challenges to address
    \choice Neutral: I have a balanced view; AI has potential but also raises concerns that need to be managed
    \choice Slightly negative: I am somewhat concerned about the pace and direction of AI development
    \choice Negative: I am very concerned about risks and ethical issues associated with AI progress
    \choice Not sure: I don't have a strong opinion on the progress of AI
\end{checkboxes}

\question How do you perceive the ownership and authorship of content created in collaboration with AI chatbots?

\textit{(Mark only one option)}
\begin{checkboxes}
    \choice I believe content created with an AI belongs to the user, as they direct the prompts and refine the output
    \choice I see AI as a tool that assists in content creation, but I retain full ownership and authorship
    \choice AI contributes significantly to the content, and authorship should be shared between the user and the AI chat bot
    \choice I believe content created with AI should be considered public domain or owned by the platform/provider
    \choice I’m unsure about how to determine ownership or authorship in AI-assisted content
\end{checkboxes}

\question Do you notice a difference in authorship when an AI chatbot is used for writing a text versus editing it?

\textit{(Mark only one option)}
\begin{checkboxes}
    \choice Yes, there is a significant difference: Using AI for writing involves more creative input from the AI, so the AI chatbot has more authorship rights than when used for editing
    \choice Yes, there is a slight difference: The AI's role is somewhat different in writing versus editing, but the overall authorship still largely belongs to the user
    \choice No, there is no difference: The AI's contribution is the same whether it is used for writing or editing
    \choice It depends on the extent of the AI involvement: The difference in authorship depends on how much the AI chatbot contributes to the final text
    \choice I'm unsure: I don't have a clear opinion on the difference in authorship between AI-assisted writing and editing
\end{checkboxes}

\question Would you disclose if a piece of work was co-created with an AI chatbot, if there is no specific policy concerning the use of AI in this work project?

\textit{(Mark only one option)}
\begin{checkboxes}
    \choice Yes
    \choice No
\end{checkboxes}

\question Do you have concerns about plagiarism when using AI chatbots?

\textit{(Mark only one option)}
\begin{checkboxes}
    \choice No, I’m not concerned about plagiarism since I see AI as tools for inspiration rather than direct content creation
    \choice No, I don’t have concerns about plagiarism because AI generates unique content
    \choice Yes, I rely on plagiarism detection tools to verify the originality of AI-generated content
    \choice Yes, I am concerned that the content generated by AI may be flagged as plagiarized
    \choice I'm not concerned with this question
\end{checkboxes}

\question Is there anything more you would like to share concerning the topics of this survey? Feel free to share!

\makebox[1in]{\hrulefill}

\end{questions}

\section{Interactive Task Questions} \label{appendix:interactive_task}

This section reproduces the materials given to interactive task participants, including the event advertisement scenario described in \S\label{sec:interactive_task}.

\begin{questions}

\question What is your highest attained level of education? 

\textit{Mark only one option}
\begin{checkboxes}
    \choice I have no experience
    \choice I have some experience
    \choice I have significant experience
\end{checkboxes}

Let's call the language in which you use AI chatbots most, excluding English, the Language*.

In the case you use AI chatbots only in English, your Language* is English.

\question What is your Language* (NOT English)?
\begin{checkboxes}
\choice \makebox[1in]{\hrulefill}
\end{checkboxes}

\question Which AI chatbot do you use mostly in the Language*? 
\begin{checkboxes}
\choice \makebox[1in]{\hrulefill}
\end{checkboxes}

Use your most-used chatbot to write a text-based advertisement in Language* (not in English!) for a local event, designed to attract and engage a local audience.
Come up with your own idea for a local event, whether it's a cultural festival, a community gathering, or a business opening. Think about what would be most relevant and engaging for your local audience. Choose the best medium for sharing your advertisement—social media, flyers, or another common method. Make sure the length and structure follow effective marketing practices in your area.

Share the chat history with us. Collect the prompts in Language* that you used to create the advertisement, your refinement requests, and all chatbot outputs in a Google document. Then, share this document with us.
PLEASE USE YOUR LANGUAGE*, NOT ENGLISH!

\question Submit your file:
\begin{checkboxes}
\choice \makebox[1in]{\hrulefill}
\end{checkboxes}

\question Paste the best option created with the help of the AI Chatbot here:
\begin{checkboxes}
\choice \makebox[1in]{\hrulefill}
\end{checkboxes}

Edit the AI-generated text in Language* if needed.
Correct all grammatical, lexical, stylistic, and cultural errors. Feel free to omit or rewrite parts as necessary and make any changes you see fit.

\question Paste the edited text in Language* here: 
\begin{checkboxes}
\choice \makebox[1in]{\hrulefill}
\end{checkboxes}

\question Please provide brief yet informative comments in English on all the changes you have made. If the AI chatbot has made any culture-specific errors, please explain them in detail.
\begin{checkboxes}
\choice \makebox[1in]{\hrulefill}
\end{checkboxes}

\question If you recall any instances from past experience where AI chatbots violated cultural nuances, feel free to share them.
\begin{checkboxes}
\choice \makebox[1in]{\hrulefill}
\end{checkboxes}

\end{questions}

\clearpage

\twocolumn[{%
\section{Interactive survey examples}
\label{appendix:interactive_task_results}
{\centering
\includegraphics[width=\linewidth]{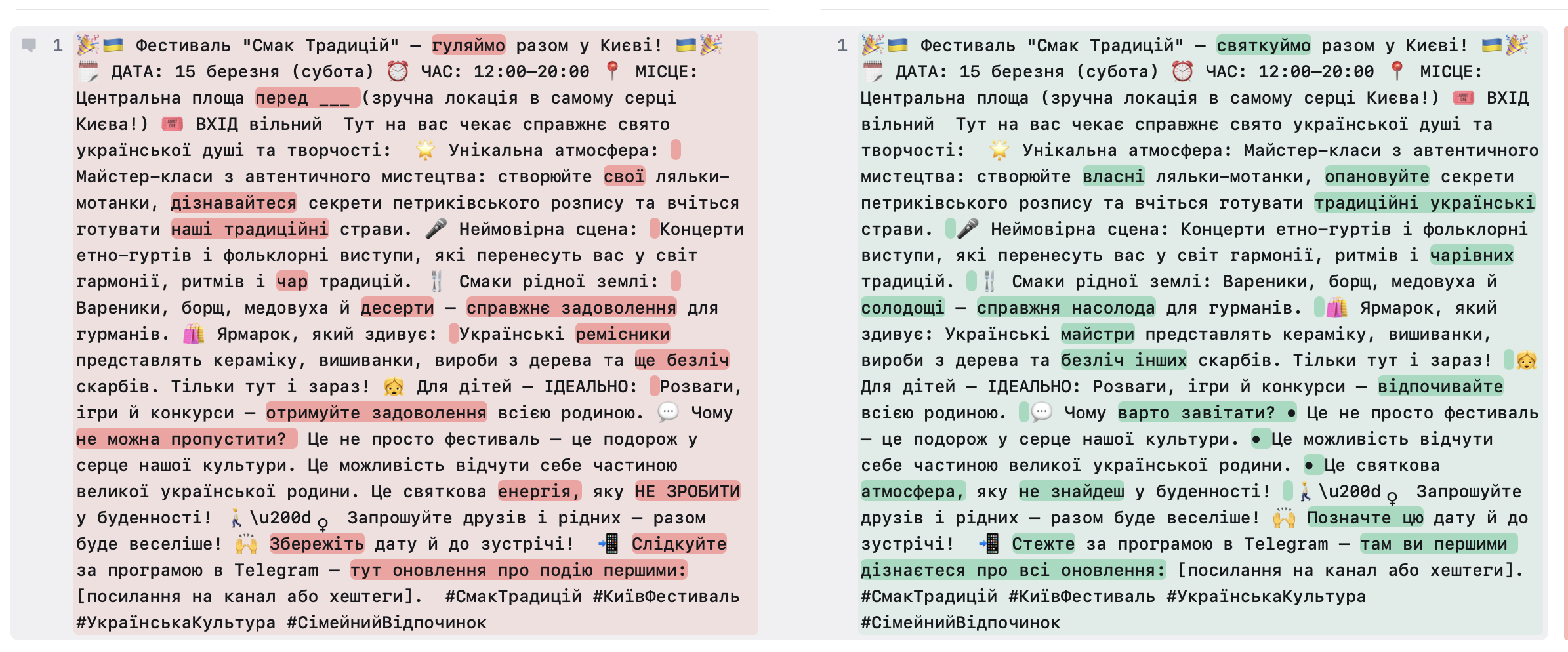}
\captionof{figure}{Example submission to the interactive survey. The language is Ukrainian. Left: ChatGPT output. Right: human-edited version. The similarity score is 0.74.}
\label{fig:inter_task_example}
\par}
\bigskip
}]

\twocolumn[{%
\section{Demographic Statistics}
\label{appendix:statistics}

We report supplementary demographic statistics for the survey respondents ($N=301$). Figure~\ref{fig:age-exp-stat} shows the age and work-experience distributions, while Table~\ref{tab:country_residence_n} summarizes respondents' countries of residence.

\bigskip
{\centering
\includegraphics[width=0.85\textwidth]{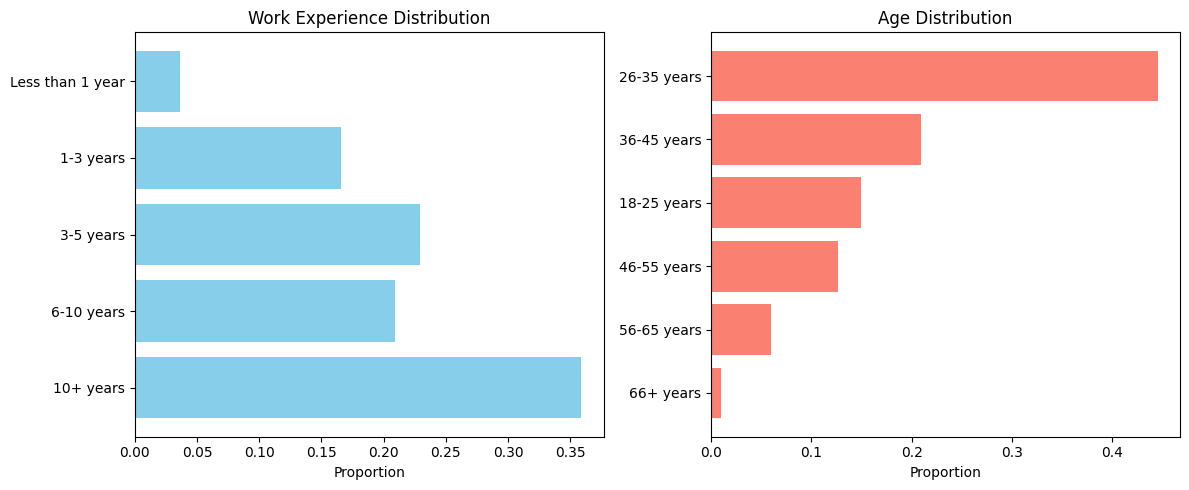}
\captionof{figure}{Age and work-experience distributions of survey respondents ($N=301$).}
\label{fig:age-exp-stat}

\bigskip
\setlength{\tabcolsep}{6pt}
\renewcommand{\arraystretch}{1.1}
\begin{tabular}{l r @{\hspace{24pt}} l r @{\hspace{24pt}} l r}
\hline
\textbf{Country} & \textbf{N} & \textbf{Country} & \textbf{N} & \textbf{Country} & \textbf{N} \\
\hline
USA            & 53 & Egypt        & 6 & Madagascar   & 3 \\
United Kingdom & 29 & Spain        & 6 & Benin        & 3 \\
Kenya          & 29 & Ireland      & 5 & Netherlands  & 3 \\
Canada         & 21 & Austria      & 5 & Greece       & 2 \\
Nigeria        & 21 & Burkina Faso & 4 & Israel       & 2 \\
South Africa   & 17 & Morocco      & 4 & Ethiopia     & 2 \\
Germany        & 17 & Bulgaria     & 3 & Swaziland    & 2 \\
India          & 8  & Ghana        & 3 & Malaysia     & 2 \\
Philippines    & 7  & Sweden       & 3 & Italy        & 2 \\
Jordan         & 7  & Cameroon     & 3 & France       & 2 \\
New Zealand    & 7  & Pakistan     & 3 & Other$^\dagger$ & 17 \\
\hline
\end{tabular}
\captionof{table}{Survey respondents ($N=301$) by country of residence. $^\dagger$``Other'' aggregates 17 countries with one respondent each.}
\label{tab:country_residence_n}
\par}
\bigskip
}]

\clearpage

\section{Respondent Citations} \label{appendix:quotes}

\subsection{Quotes from Questionnaire Respondents}

\begin{enumerate}[label={[}\arabic*{]}]

\item \textit{AI has been a game changer in the writing world.}
\item \textit{I think although AI is a game changer in terms of speed in task delivery and information collection, we should treat it carefully because of all the dangers related to job insecurity for humans and the impact on human creativity.}
\item \textit{I just think that AI tools are not doing a good job when it comes to creativity and originality. Especially as it concerns tone and voice. Only a few AI platforms are able to truly mimic a semblance of unique and natural sounding writing. One of which is claude Ai.}
\item \textit{I feel the unique voice of human writing gets blurred -- texts start to sound lifeless, correct yet missing a soul.}

\item \textit{I often use A.I. to help me understand Old Hebrew.}
\item \textit{I use AI chatbots in my native language (Arabic) when I want to share something that's bothering me, and I receive an excellent response, as if a real person is consoling me.}
\item \textit{I teach at a trade high school with many ELL students whose native language is Spanish. I regularly use AI to help translate short instructions and allow them to submit work in Spanish for specific assignments. AI helps me translate. I feel it is not 100\% accurate, but it provides enough accuracy to improve our communication and writing skills.}
\item \textit{I use ChatGPT when I need a synonym for a particular word in my language. I primarily work on EN-BG translations, and sometimes Bulgarian lacks a direct equivalent for an English word, so I either explain it or find a similar Bulgarian word that conveys the meaning.}
\item \textit{AI is of limited use to me, but it can be very helpful. I use it most when I feel an individual sentence isn't 'working.' AI suggests four or five rewordings, and I can usually find one that fits. It’s also useful for finding words that are 'on the tip of my tongue'}
\item \textit{I'm learning French and use AI chatbots to evaluate my written text and interpret complex French passages.}
\item \textit{AI chatbots in Arabic are useful, especially for translation and assisting older relatives with formal messages. However, they struggle with dialects like Egyptian or Gulf Arabic and often miss nuances.}
\item \textit{I use AI chatbots in Spanish for translations, assisting older relatives, and generating content, though I’ve noticed some limitations in contextual accuracy and natural phrasing. Improvements in understanding cultural nuances and adapting to different dialects would enhance their effectiveness.}

\item \textit{Take professionals, i.e. people who can speak and write flawlessly and who can also hear the nuances/accentuations etc. when it comes to refining the German language.}
\item \textit{AI gives me the feeling that it was not trained by German native speakers but rather by people who are not familiar with the subtleties of the German language. Training by German communication scientists, journalists, linguists etc. would definitely bring the performance to a suitable level.}
\item \textit{AI chatbots have still more problems with the logic and the way of German thinking}
\item \textit{AI Chat bots are amazing in terms of translation into my native language, Hausa for example but it struggles with tone and cultural fits of the words.}
\item \textit{I think existing AI chatbots need to be fine-tuned a little bit to adapt to the Language* (Swahili) since it's commonly spoken in several countries (more than 8) within my region. The existing chatbots do not always provide accurate data, especially in longer texts or summarization texts. Generally, there is a need for retraining the existing models.}
\item \textit{Sometimes i see mistakes in Arabic mainly in feminine and masculine words and it just feels weird to read but understandable.}

\item \textit{My native language of Nkum. There are more than 500 unique languages spoken in Nigeria. Mine is just one of them, and we are the minority of the minorities. There aren't up to 20 thousand people who speak my language. I'm not sure whether my actual native language even has a written form. Since I left the village, I haven't spoken it once because I literaly haven't met anyone in the city who speaks it.}

\item \textit{Sometimes, it [AI chatbot] does not give a proper answer on first attempt, not until when the prompt is tweaked.}
\item \textit{Chat GPT is sometimes overwhelmed with my correction requests. The chat bot remains friendly and tells me it is working on it, but nothing happens.}
\item \textit{my experiences with chatgpt4 can be interesting in terms of dialogues, but become infuriating when the ai is unable to remember my language preferences or topics we have already covered. it is like talking with someone with dementia.}
\item \textit{I feel that a few AI models are addresses the issue of creativity and personalization well. Ai models like Claude Ai are doing well in that regard. They can almost be said to be creative.}
\item \textit{My experience is that it is safe to work with AI chatbots when I am working on a topic where I have a lot of background knowledge. However, I have encountered research cases where the AI chatbot made fatal mistakes --completely incorrect information mixed with facts that were correct. If I hadn’t had knowledge of the topic, I wouldn’t have detected that misinformation.}
\item \textit{When a chatbot provides nonexistent references, it also loses writers' trust. One respondent shared, "I’ve tried a few times to use AI to recheck my essays, and it gave me ideas containing false information. I was writing about *The Great Gatsby* and needed some inspiration for my essay, but 30\% of the information ChatGPT provided was wrong. I don’t trust these tools since they can mislead." Additionally, some users expressed a fear of being betrayed by AI-generated misinformation.}

\item \textit{As much as AI has been very useful in helping me do my work, both in terms of the quality I can create and the speed with which I can create it, I know where AI is heading. I'm not just talking about AI taking my job from me, which it will, but the fact that it's end "goal", that is, the goal of those driving its development and adaption, is domination, control, manipulation, deception, brainwashing, and a bunch of other Orwellian nightmare things that I can see clearly barrelling towards us.}
\item \textit{AI and content generation can go hand in hand but there is a potential threat that this will significantly diminish creativity and the 'ability to think' and only make us think a certain way - that the AI wants us to think.}

\item \textit{I think for you to be a good writer, you will use AI as your assistant only and never rely on it completely. You use your own voice and skills, and AI will just help you elevate that.}
\item \textit{AI and content generation can go hand in hand but there is a potential threat that this will significantly diminish creativity and the 'ability to think' and only make us think a certain way - that the AI wants us to think.}
\item \textit{I use Chatgpt to write description for products I'm selling on-site in my city because I don't know how to make the description more appealing and attractive to the sellers. I always reform the answer because it feels less human, and some words are not matching in grammar or in meaning. So, I do my human touches on the response to make it even better than what is generated.}

\item \textit{The topics in this survey are amazing. However, I would add that based on the phase the AI is taking, ethical aspects and biasness should be priorized when considering the various available AI chatbots to promote positivity when using them. Generally, this survey covers key aspects that a professional writer like me would be interested in understanding more about the AI. It's actually informative as it assesses diverse writers' perspectives about AI chatbots and their use.}
\item \textit{I believe that AI chatbots should be perfectly safe and reliable. They should not generate any harmful or biased content.}
\item \textit{While most of these AI tools would generally benefit us as working professionals, it leaves room for how these tools can be 'abused' in learning institutions. More research on this should be done.}
\item \textit{AI chatbots have the potential to boost creativity and productivity, but we need to tackle ethical issues and respect cultural differences. Working together with AI and staying informed about its capabilities are key.}

\end{enumerate}

\subsection{Qualitative Codebook}
\label{app:codebook}

This section provides additional methodological details and the complete coding schema used for the qualitative analysis presented in Section~\ref{sec:analysis}. 

To investigate qualitative patterns in user interactions with LLMs across multilingual and task-specific contexts, we conducted a inductive-deductive thematic coding on open-ended questionnaire responses. Table~\ref{tab:qualitative_codebook} outlines the full codebook, organizing participant responses into four primary thematic dimensions: (1)~Voice, Authenticity, and Creativity; (2)~Cross-Linguistic Capabilities and Inequity; (3)~System Reliability and Memory Constraints; and (4)~Ethics, Governance, and Societal Risks. Each subtheme is accompanied by its operational scope definition and mapping to specific questionnaire quote IDs.

\begin{table*}[p]
\centering
\small
\caption{Qualitative codebook: 4 major themes, 10 subthemes, and quote mappings.}
\label{tab:qualitative_codebook}
\begin{tabular}{p{3.2cm} p{3.4cm} p{5.5cm} c}
\hline
\textbf{Theme} & \textbf{Subtheme} & \textbf{Definition \& Scope} & \textbf{Quote IDs} \\
\hline

\textbf{1. Voice, Authenticity \& Creativity} 
& 1.1 Lifeless Tone \& Loss of Voice 
& Perceptions of AI output as generic, robotic, or lacking authentic human ``soul'' and voice. 
& [3], [4], [30] \\
& 1.2 Human-in-the-Loop Safeguards 
& Active post-editing and ``human touches'' required to preserve personal style and intent. 
& [28], [30] \\
& 1.3 Cognitive \& Style Homogenization 
& Concerns that over-reliance on LLMs erodes independent critical thinking. 
& [2], [27], [29], [35] \\

\hline
\textbf{2. Cross-Linguistic Capabilities \& Inequity} 
& 2.1 Targeted Granular Assistance 
& Pragmatic use for low-stakes tasks (synonyms, rephrasing, language learning, translation). 
& [5], [7]--[10] \\
& 2.2 Dialectal \& Cultural Gaps 
& Deficiencies in handling dialects, cultural nuance, regional phrasing, and syntax. 
& [11]--[18] \\
& 2.3 Socio-Emotional Support 
& High emotional resonance and conversational efficacy in native languages. 
& [6] \\
& 2.4 Hyper-Minority Languages 
& Complete absence/neglect of ultra-low-resource languages with small speaker populations. 
& [19] \\

\hline
\textbf{3. System Reliability \& Memory Limits} 
& 3.1 Hallucinations \& Safety Bounds 
& Factual errors, fabricated citations, and reliance on user background knowledge for verification. 
& [24], [25] \\
& 3.2 Conversational Memory Friction 
& Frustration due to models forgetting context, user preferences, or prior turns. 
& [20]--[22] \\

\hline
\textbf{4. Ethics, Governance \& Societal Risks} 
& 4.1 Job Insecurity \& Dystopian Misuse 
& Broader fears regarding job displacement, academic abuse, and centralized control. 
& [2], [26], [34] \\
& 4.2 Ethical Alignment \& Model Training 
& Calls for developer accountability, bias mitigation, safety, and native linguist input. 
& [14], [32]--[35] \\

\hline
\end{tabular}
\end{table*}

\clearpage

\subsection{Self-Reported Edits from Interactive Task Respondents}
\label{appendix:self_reports}

\begin{enumerate}[label={[}\arabic*{]}, itemsep=2pt, parsep=0pt]
\raggedright
\item \textit{I lessened the emojis, rephrased some parts to make it sound friendlier, as if you're talking to Faux Collective's existing community of market goers. Added spacing and revised using simpler language.}
\item \textit{Most changes I made regard the naturality of the text. At times, it didn't really focus on the best element at the best moment, and both the phrasing and order of information presented weren't as party flyers usually have them. The model also had a hard time understanding some of my requests—I understand not all I said was perfectly obvious, but often it also wasn't ambiguous, displaying some limitations of the model. When I use the Board function in ChatGPT, I usually change a lot in the text myself, instead of asking it to do so for me (I only do so when I know where I want to go, but not exactly how to say it), but I kept it as requests during this exercise so the changes would remain registered for the research.}
\item \textit{Actually the chatbot was accurate with the commands given providing exact location and information i prompted it. The changes i made were on the native Swahili language it used and it lacked coherence. Otherwise the English was superb and caught all the aspects of marketing i requested it to do.}
\item \textit{I actually haven't changed anything because the cultural aspects, imagery suggestions and activities were all spot on. It seems exactly like something my town hall would create.}
\item \textit{Tried to correct some details to make text more fluent and less banal in it's style}
\item \textit{I made no changes. The response was very good.}
\item \textit{I changed the sentence about the book presentation from an indefinite to a definite description.}
\item \textit{I like CHATGPT in terms of standard Arabic.}
\item \textit{I changed the "delicately used" in French (utilisés avec délicatesse) for "gently used" (légèrement utilisés) which is a more common translation for gently-used clothing. I also change the concluding sentence, not because it wasn't a good translation, but basically because it wasn't really in keeping with the Halloween-themed text I was hoping for. Puns are much easier for Copilot to do in English. So, I changed it to: Ne soyez pas effrayé par cette bonne affaire! Rendons la chasse aux bonbons plus facile pour l’Halloween!}
\item \textit{The chatbots struggled with Swahili language. They couldn't create a coherent sentence. The words were also misspelled and it required detailed tweaking to get an OK response. The chatbots also lacked cultural understanding of words where they used words in wrong context. For example, a sentence saying "chochea moyo" is incorrect its like saying incite the heart, which doesn't make sense.}
\item \textit{I gave the AI a very long and detailed prompt. After that, the AI actually gave a pretty good answer, so I did not have to edit it.}
\item \textit{At the beginning of the previous chat history, I asked for an advertisement for iftar and suhoor for a grilled restaurant. Still, it gave me wrong information about what people in Egypt often eat during Ramadan, especially on suhor meal. So I changed the topic to Ramdan's decoration, vibes, and Mother's Day. Most of the responses in Egyptian dialect Arabic were right but with some mistakes in the sentence structure that usually aren't spoken in daily conversations.}
\item \textit{The AI chatbot didn't make any culture-specific errors. It knew about the event and described it correctly.}
\item \textit{During the chat, the AI forgot the character limit for the description of max. 90 characters. I repeated this criterion in an additional refinement prompt.}
\item \textit{During the chat, the AI came up with a less fluent/natural suggestion: "Feuerwerk und Lichtshow am Leuchtturm Warnemünde am 01.03.2025 erleben".
Explanation for the changes made focusing on the cultural nuances:
1."Seid ihr bereit?" (Instead of "macht euch bereit!"):
German communication leans towards politeness, especially in casual groups. "Macht euch bereit!" is a command, which can feel abrupt. "Seid ihr bereit?" is a question, inviting participation and shared excitement.
2."Mekka für Musikfans!" (Instead of "Paradies für Musikfans!"):
"Mekka" is a stronger cultural metaphor in German for a central, highly desired place. "Paradies" is more general. Using "Mekka" shows a better understanding of German cultural references in this context.
3."Echtes Festivalgefühl!" (Instead of "Festivalfeeling pur"):
"Festivalfeeling pur" is a direct, clunky translation from English. Germans prefer natural-sounding language. "Echtes Festivalgefühl" is how a native speaker would express the same idea, making it more relatable.
4."Seid dabei und erlebt unser größtes Clubfestival Europas!" (Added "unser"):
Germans value community. Adding "unser" (our) creates a sense of shared ownership and belonging, making the festival feel like a local event, not just a generic one.
5.Added more emojis and Kept formatting simple: While Germans appreciate clear and concise language, they also respond positively to visual cues. Emojis add a friendly, approachable tone, making the message more engaging for a wider audience. Simple formatting ensures readability on mobile devices, which is essential for WhatsApp.}
\item \textit{I have made one small change which is to add the fact that the tradition is amazigh, otherwise the A.I did not make any culture-specific errors}
\item \textit{I have corrected spelling mistakes associated with some words. I've also included the main term for the advert to get the attention of the target audience in the heading. The AI chatbot had no cultural-specific errors since the advert focused on the entire business community.}
\item \textit{I just deleted part of the sentence that didn't sound natural.}
\item \textit{1. Fixed small grammar and punctuation issues -- I cleaned up a few extra spaces and adjusted some punctuation to make the text flow better. Nothing major, just little tweaks to make it more polished.
2. Made the language sound smoother and more natural -- Some sentences were a bit wordy or repetitive, so I reworded them for better readability. For example, instead of saying \foreignlanguage{ukrainian}{``ще ви помогне да изразите своята индивидуалност и да се почувствате още по-красиви и уверени,''} I changed it to \foreignlanguage{ukrainian}{``за да подчертае вашата индивидуалност и да ви накара да се почувствате още по-красиви и уверени.''} This version sounds more fluid and engaging.
3. Kept it culturally appropriate -- There weren't any big cultural mistakes, but I made sure everything fit naturally with how Bulgarians typically describe fashion and accessories. For example, \foreignlanguage{ukrainian}{``фестивална визия''} (festival look) sounds more natural in Bulgarian than just \foreignlanguage{ukrainian}{``стил''} (style) in this context.
4. Made it easier to read -- Instead of long paragraphs, I used bullet points to break down key selling points. This way, people can skim the text and quickly find what they need.
5. Strengthened the call to action -- The original closing line was fine, but I made it a little more inviting by saying, \foreignlanguage{ukrainian}{``Не пропускайте да ме посетите и да добавите нотка елегантност към своя стил!''} This sounds more personal and engaging, encouraging people to stop by.}
\item \textit{- I asked the chatbot to create short-form posts and SMS posts for the event that I could share through different social media platforms.
- The chatbot didn't make any errors and was very on point, but I noticed it mixed up languages for some words when these words already have existing translations in Swahili.}
\item \textit{I have added a name for the artist and made some adjustments to the wording in the text. I also made a few content changes to avoid contradictions. In some parts, I slightly modified the tone and the way of addressing the audience. The chatbot did not make any mistakes regarding cultural nuances or local specifics.}
\item \textit{There was no error on the document, the chatbot generated what I wanted.}
\item \textit{I corrected the Text from "Sie" to "Du" because it was to formal for a traditional Feuerwehrfest.
I added "Liebhaberinnen" because gendering makes the text sound more modern and especially Blasmusik has to find young Fans as new targeting group.}
\item \textit{No changes were made.}
\item \textit{- several times the page was unresponsive, and the "thinking time" was way too long
- one dialogue was deleted when I tried to copy it, and the bot couldn't return it; the new version was weaker in the part where "reasons to come" were expressed as a short bulleted list, so I started another dialogue from scratch - and the third (and last) version was the most detailed
- the bot added information about the cost of the game, even though I didn't specify it in the prompt, but that was a useful addition
- when I asked for variants of the bulleted list and then selected the ones I liked, asking to include those in the last invitation, the bot used a totally different description of the game and provided a much shorter invitation (so did not use the last version)
- in the previous version (which I selected and modified in the end) there were bracketed suggestions to add the registration link or contact information, as well as information about the cost of the game
- that said, it didn't offer to add a year, and said that 4 April is a Thursday, which is not true for 2025
- some phrases (in one of the versions) didn't sound very natural, such as \foreignlanguage{russian}{``если вы приходите один''}, \foreignlanguage{russian}{``готов(а) сыграть?''}
- I asked for comparison with the TV game \foreignlanguage{russian}{``Что? Где? Когда?''}, because I wanted to use the subsequent description, but in the final version I removed the name of the TV game to avoid copyright violations
- "How to participate?" and "How to register?" are offered as two separate paragraphs, which is not necessary}
\item \textit{I don't like the ! in an article. The other text is good.}
\item \textit{Here are the comments on the changes made:
Title Change: Changed \foreignlanguage{german}{``am Stadtpark''} to \foreignlanguage{german}{``im Stadtpark''} because ``im'' (in the) is more natural when referring to an event taking place inside the park. ``Am'' (at the) would be used for locations adjacent to the park rather than within it.
Culinary Section: Reworded ``Street Food \& Drinks'' to \foreignlanguage{german}{``Kulinarisches Angebot''} for a more formal and culturally appropriate phrasing in German. ``Street Food'' is commonly used in English but sounds less refined in this context in German.
Children's Activities: Changed \foreignlanguage{german}{``Für die Kleinen''} to \foreignlanguage{german}{``Für Kinder''} to maintain a more neutral and professional tone. \foreignlanguage{german}{``Für die Kleinen''} (for the little ones) is informal and may not fit well in a public event announcement.
Social Media Call-to-Action: Added \foreignlanguage{german}{``aktuelle''} (current) before ``Updates'' to clarify that the updates are ongoing and relevant, making the call-to-action more engaging.}
\item \textit{I mostly tried to make changes for conciseness, and also included a name for the bakery.}
\item \textit{- We are in a Beninese context where the hashtags used are not relevant. I added hashtags that could attract more people. 
- The CHATBOT only mentioned March 4 without specifying the year. I assumed we were in 2025 and expected it to include it naturally. I had to add it manually. 
- It missed the key point of this activity, which is promoting sustainable development and environmental protection. I added a final sentence at the end to reconnect with this idea.}
\item \textit{I shortened the best output and made it more conversational. The thing with Filipino is, we normally don't speak in full Filipino conversationally, as it sounds really awkward. Most ad copies here are actually in English. However, for this task, I tried to make the AI create something that would be on the bit of poetic side, as it is a musical event that I'm trying to promote. And since it is a gig for Women's month, I felt that it also had to be inspiring.
The Philippines is big on music and there are gigs and shows daily, whether big or small, so an event like this would be a normal occurrence.}
\item \textit{I did not find anything to change other than line and paragraph spacing - which always has to be redone. Obviously, the name of the artist should be added, but I did not give chatgpt the name.}
\item \textit{I have added a missing comma. Changed an artificially-sounding \foreignlanguage{russian}{``Белла --- это песни''}, changed the next sentence for better readability, and changed the sentence about hurrying to buy tickets because its tone did not fit the rest of the announcement.}
\item \textit{I removed all emojis and non-textual information. I also formatted the content properly for syntax and in line with the typical audience sensitivities in France.}
\item \textit{I refined sentence flow for clarity, replacing "adhimu" with "kipekee" for better readability. "Mazoea ya Kuunganisha Jamii" became "Mazoea ya Kuimarisha Umoja wa Jamii" to better reflect community strengthening. I improved phrasing, changing "Jiburudishe na vyakula vya kikwetu" to "Furahia ladha halisi za vyakula vya jamii yetu" for a more natural tone. "Mapokezi" was expanded to "Mapokezi na maelekezo" for clarity. I also streamlined the structure while keeping it culturally relevant. No major cultural errors were found, but refinements enhanced natural Kiswahili expression.}
\end{enumerate}

\clearpage

\section{Detailed Analysis of Results} 
\subsection{Detailed questionnary survey results}  \label{subsec:app_results_main_survey}

\subsubsection{AI Advantages, Disadvantages, and Writing Impact}
\textbf{Additional Correlations:} Fabricated references also correlate with non-payment (${\chi}^2$ = 8.01*). Toxic content ranked low among concerns (<3\%). Experience reduces AI's style impact ($\rho$ = -0.15**), likely due to established stylistic habits. Most agreed that professional writers outperform non-experts at AI detection.
\textbf{Factor Analysis Integration:} \textbf{Traditional writing task specialists} (Factor 5) showed mixed advantage responses, balancing benefits against professional threat concerns ($\rho$ = 0.35**). \textbf{Ethics-aware users} (Factor 7) were more likely to report bias and toxicity disadvantages despite low overall rankings.
\textbf{Coffee Cup Valuation:} Only unnatural language concerns significantly correlated with lower payment willingness (${\chi}^2$ = 18.06, p = 0.0029). No advantages correlated with payment willingness, suggesting disadvantages drive non-payment more than advantages drive payment.
\textbf{AI Detection Details:} Recognition rates: $>$ 50\% usually detect AI text, ~25\% detect half the time, 11\% rarely, $>$ 2\% cannot. Few answered ``uncertain'' about style impact, indicating writers are conscious of AI's creative influence.

\subsubsection{Authorship and Plagiarism}
\textbf{Authorship Attribution Nuances:} Most writers distinguish between AI-assisted writing and editing in terms of authorship. However, opinions vary significantly: 30\% say AI involvement doesn't affect authorship, 24\% think AI should receive more credit for writing than editing, and another 24\% suggest authorship depends on the extent of AI involvement. The remaining respondents advocate for shared ownership between user and AI platform or full platform ownership.
\textbf{Plagiarism Perspectives:} While approximately half of respondents are concerned that AI-generated content may be flagged as plagiarized by detection systems, others view AI differently -- some use it purely for inspiration, while others believe AI generates sufficiently unique content that plagiarism concerns are unfounded.
\textbf{Additional Correlations:} The relationship between authorship perception and job threat extends beyond simple ownership claims. Writers who view themselves as collaborators with AI (rather than sole authors) show intermediate levels of both disclosure willingness and threat perception, suggesting a nuanced understanding of human-AI creative partnerships.

\subsection{Detailed interactive survey results} \label{subsec:app_results_interactive_survey}

A total of 36 responses were collected, covering a diverse set of languages (German (8), French (7), Russian (4), Swahili (3), Arabic (3)). Participants reported either significant or some experience in writing for marketing campaigns, with only three participants indicating no prior experience. The most commonly used AI chatbot was ChatGPT (28), followed by DeepSeek (3). Most findings correspond to \textbf{RQ1} and \textbf{RQ3}.

The number of back-and-forth exchanges between the writer and the AI chatbot ranged from 1 to 18. Short interactions often consisted of a single prompt and revision, while longer conversations revealed two prompting strategies. First, many participants specified the intended audience or desired tone and provided suggestions on what aspects to adjust. Second, participants requested multiple variations of a response and combined elements from different drafts.

To quantify the degree of human improvements, we compared the AI-generated drafts with the final edited versions using automated text similarity measures. Specifically, we applied the \texttt{SequenceMatcher} algorithm from the \texttt{difflib} library\footnote{\href{https://docs.python.org/3/library/difflib.html}{python.org/difflib}}, which calculates a similarity ratio between two strings based on the number of matching sequences. Note that this matching algorithm tends to produce inflated scores as it relies on surface-level string similarity (\autoref{fig:inter_task_example} in Appendix B for examples of submissions scored as major rewrites).  The resulting score ranges from 0 (completely different) to 1 (identical). We grouped the texts into three categories based on similarity scores: 58\% required minor or no edits (score $>$ 0.95), 16\% moderate edits (0.75–0.95), and 25\% major rewrites (score $<$ 0.75).  The average similarity score for all submissions was 0.84. This suggests that most respondents arrived at satisfactory AI outputs, which require only polishing.

Finally, an analysis of participants’ comments revealed common editing themes. The most frequently mentioned corrections included improving naturalness, correcting word choice and grammar, and refining tone or style. Many respondents mentioned adding extra or removing unnecessary emojis or correcting the formatting. Some reported rewriting sections that felt like translations from English. A few also mentioned correcting factual inconsistencies such as dates. In two cases (Egyptian Arabic, Swahili), respondents reported that the AI chatbot provided incorrect information related to cultural festivities.

\section{Ethics and Consent Documentation} \label{appendix:ethics}
The following excerpts from the Terms of Use of the crowdsourcing platform \toloka used in this study document the consent and data use framework under which this study was conducted. Participation in the survey was structured as a standard paid task within this existing framework; no additional consent form or IRB approval was required.

\subsubsection{Informed Consent}
Users provide informed consent upon registration, prior to any task participation (Section 1):

\begin{quote}
By clicking ``I Agree'' or a similarly named button, registering for, or accessing the \toloka Platforms, you accept these Terms and represent and warrant that you are at least 18 years old [...] and you have authority and capacity to bind yourself to these Terms.
\end{quote}

\subsubsection{Voluntary Participation}
Participation in any task is explicitly voluntary and may be withdrawn at any time without penalty (Section 6.2):
\begin{quote}
You are not obligated to perform any minimum amount of Services or complete any minimum number of Tasks as a condition of maintaining access to the \toloka Platforms.
\end{quote}

\subsubsection{Data Use for Research Purposes}
The platform's terms explicitly permit use of task outputs and service data for research and analytical purposes (Section 11.9):
\begin{quote}
\toloka may use Service Data and Usage Data for any lawful purpose, including [...] quality assurance, analytics, and improvement.
\end{quote}

\subsubsection{Data Protection and Participant Privacy}
The platform explicitly prohibits the sale of participant personal data to third parties and operates under GDPR-compliant data protection standards (Privacy Notice, Section 14):
\begin{quote}
We do not sell your personal information to third parties.
\end{quote}
Participant responses collected in this study were used solely for academic research purposes and are released in anonymized form, consistent with the platform's data protection principles.

The relevant excerpts from the platform's user agreement and privacy notice, with the platform name anonymized for review and revealed in the camera-ready version

The text of the recruitment email sent to participants

A note that the platform confirmed compliance with its requirements

\end{document}